\documentclass{article}
\PassOptionsToPackage{table}{xcolor}
\usepackage[submission]{colm2026_conference}

\usepackage{xcolor}
\usepackage{hyperref}
\usepackage{url}
\usepackage{booktabs}
\usepackage{lineno}
\usepackage[most]{tcolorbox}
\usepackage{caption}
\tcbuselibrary{skins}

\usepackage{amsmath,amsfonts,bm}

\def\eqref#1{equation~\ref{#1}}

\def\1{\bm{1}}

\DeclareMathAlphabet{\mathsfit}{\encodingdefault}{\sfdefault}{m}{sl}
\SetMathAlphabet{\mathsfit}{bold}{\encodingdefault}{\sfdefault}{bx}{n}

\usepackage{graphicx}
\usepackage{tabularx}
\usepackage{amsmath}
\usepackage[capitalize]{cleveref}
\usepackage{arydshln}
\usepackage{mathtools}
\usepackage{multirow}
\usepackage{lmodern}
\usepackage{setspace}

\newif\ifshowcomments
\showcommentsfalse

\definecolor{absgray}{RGB}{242,243,245}
\definecolor{metablue}{RGB}{0,102,204}

\usepackage{comment}
\usepackage{array}
\usepackage{multicol}
\usepackage{makecell}
\usepackage{rotating}
\usepackage{lipsum}
\usepackage{enumitem}

\newcommand{\algname}{{\texttt{{ReMEMBER}}}}
\newcolumntype{L}[1]{>{\raggedright\let\newline\\\arraybackslash\hspace{0pt}}m{#1}}
\newcolumntype{X}[1]{>{\centering\let\newline\\\arraybackslash\hspace{0pt}}p{#1}}
\newcolumntype{C}[1]{>{\centering\let\newline\\\arraybackslash\hspace{0pt}}m{#1}}
\newcommand{\brk}{\discretionary{}{}{}}
\newcommand{\idlink}[2]{\href{#1}{\texttt{#2}}}

\usepackage{wrapfig}

\usepackage{pgf}

\definecolor{ChatColor}{RGB}{80,130,200}
\definecolor{MeetColor}{RGB}{80,160,120}
\definecolor{AllColor}{RGB}{150,110,190}

\colorlet{oraclerow}{gray!12}
\colorlet{proposedrow}{teal!10}

\newtcolorbox{promptfigbox}[2][]{
  enhanced,
  width=\linewidth,
  colback=gray!3,
  colframe=gray!35,
  colbacktitle=gray!15,
  coltitle=black,
  title={#2},
  fonttitle=\bfseries\footnotesize,
  fontupper=\scriptsize,
  before upper={\linespread{1.00}\selectfont\raggedright},
  boxrule=0.35pt,
  arc=3pt,
  left=5pt,
  right=5pt,
  top=4pt,
  bottom=5pt,
  boxsep=1pt,
  titlerule=0pt,
  #1
}

\newlist{promptenum}{enumerate}{1}
\setlist[promptenum]{
  label=\arabic*.,
  leftmargin=1.2em,
  labelsep=0.4em,
  itemsep=2pt,
  topsep=2pt,
  parsep=0pt,
  partopsep=0pt
}

\newlist{promptitems}{itemize}{1}
\setlist[promptitems]{
  label={\raisebox{0.15ex}{\scriptsize$\bullet$}},
  leftmargin=1.3em,
  labelsep=0.45em,
  itemsep=1pt,
  topsep=1pt,
  parsep=0pt,
  partopsep=0pt
}

\newcommand{\customabstractpage}{
\begin{tcolorbox}[
    enhanced,
    colback=absgray,
    colframe=absgray,
    boxrule=0pt,
    arc=8pt,
    left=3mm,
    right=3mm,
    top=3mm,
    bottom=3mm
]

{\Large\bfseries
Don't Scroll Back: Missing-Evidence Memory for Streaming Dialogue Summarization
\par}

\vspace{3mm}

Hyangsuk Min, Hwanjun Song\par

\vspace{1mm}

KAIST\par

\vspace{4mm}

\noindent
Users of modern platforms repeatedly need summaries of recent dialogue, but the window rarely contains enough context to be interpreted on its own. We formalize this setting as {streaming dialogue summarization}, where a system must summarize a current window using selective memory from an unbounded history under a fixed budget. We show that the central challenge is not how much history is accessed, but whether memory recovers the evidence that the current window presupposes. We construct a benchmark and evaluation protocol that separately assesses whether memory contains gap-resolving evidence and whether the generated summary reflects it. We propose \algname{}, a missing-evidence memory framework that conditions retrieval on unresolved window dependencies and refines retrieved chunks into evidence-dense memory under a fixed budget. Experiments on dialogues with histories up to 160K tokens show that \algname{} improves memory recall and gap-resolution completeness over memory construction baselines under the same budget.
\vspace{4mm}

\noindent
\begin{minipage}[t]{0.65\textwidth}
{\small
\textbf{Date:} Aug 10, 2026 \par
\textbf{Correspondence:} Hwanjun Song at {\color{metablue}\href{mailto:songhwanjun@kaist.ac.kr}{songhwanjun@kaist.ac.kr}} \par
\textbf{First Author:} Hyangsuk Min at {\color{metablue}\href{mailto:hyangsuk.min@kaist.ac.kr}{hyangsuk.min@kaist.ac.kr}} \par
}
\end{minipage}
\hfill
\begin{minipage}[t]{0.27\textwidth}
\vspace*{-0.1cm}
\raggedleft
\includegraphics[width=1.0\linewidth]{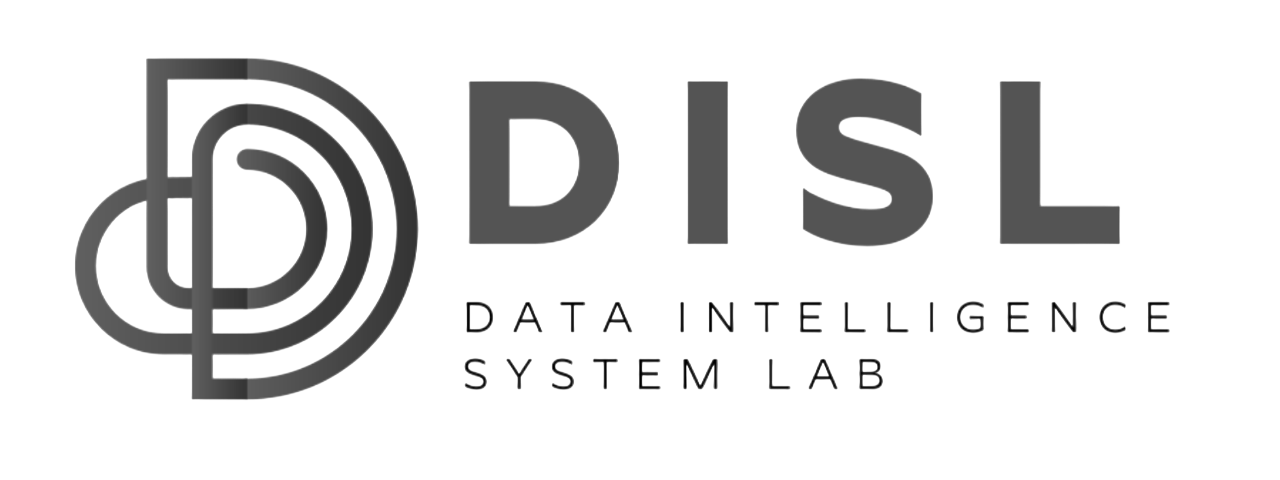}
\end{minipage}

\end{tcolorbox}
}

\begin{document}

\thispagestyle{empty}

\vspace*{-1.3cm}
\customabstractpage

\section{Introduction}
\label{sec:introduction}
Modern communication platforms, such as message apps (e.g., WhatsApp, WeChat) and collaborative workspaces (e.g., Slack, MS Teams), generate continuous streams of multi-turn dialogues \citep{yi2025survey, kirstein2025cads, deshpande2025multichallenge}. Users rarely seek a global summary, and instead repeatedly need summaries of recent segments \citep{wang2024dialogue, ghebriout2025quartz, wu2025incremental} to catch up on recent discussions or decisions. We refer to this as \textit{streaming dialogue summarization}. The core challenge is that the current window is seldom self-contained. Speakers rely on shared history, leaving pronouns without antecedents, entities without attributes, and decisions without rationales. Summarizing the window alone yields ungrounded fragments, while resolving these dependencies requires retrieving evidence from earlier turns under a strict budget.

Existing dialogue summarization, however, is largely designed under a \emph{closed}-dialogue assumption \citep{zhu2025factual, jin2025reasoning}, where the full conversation is available and a single global summary is produced. To scale to long dialogues, prior methods adopt either \emph{incremental} summarization \citep{wu2025incremental, wang2025recursively}, which updates a running summary as new turns arrive, or \emph{hierarchical} summarization \citep{kim2025nexussum, li2025hierarchical, ou2025context}, which summarizes small segments independently and then merges them. While effective for constructing a global dialogue summary, they are misaligned with streaming scenarios. The former attenuates earlier context through repeated compression, while the latter favors global abstraction and discards the fine-grained evidence needed to interpret a specific window. Neither preserves the prior context that resolves dependencies in the current window.

\definecolor{evidencegreen}{HTML}{1B6E3A}
\begin{figure*}[t!]
    \centering
    \includegraphics[width=0.99\textwidth, trim=0 0 0 0, clip]{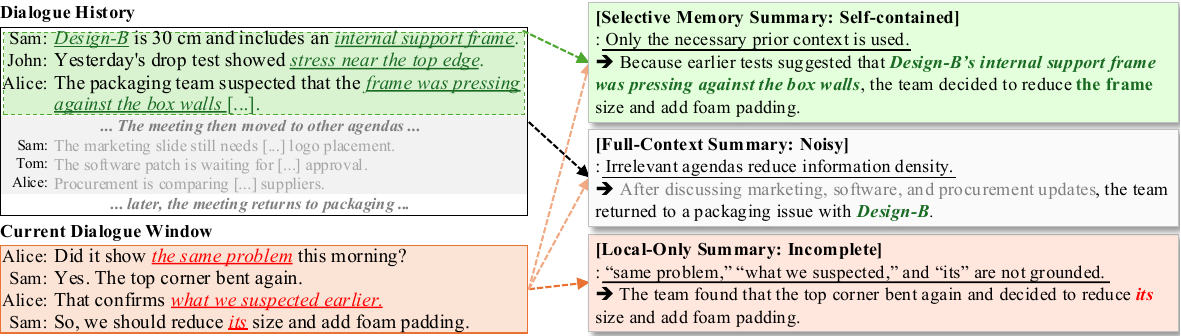}
    \caption{Context utilization in streaming dialogue summarization. \textit{Local-Only} leaves contextual dependencies unresolved ({\color{red} red}) and \textit{Full-Context} introduces irrelevant history~({\color{gray} gray}). \textit{Selective Memory} addresses both by retaining only necessary prior evidence for a self-contained summary ({\color{evidencegreen} green}).
}
    \label{fig:intro}
\end{figure*}

One might simply feed the entire dialogue history to recover the missing context. In practice, however, histories routinely span tens to hundreds of thousands of tokens \citep{lee2025realtalk}, where this approach is computationally prohibitive and unreliable due to well-known \emph{long-context} pathologies, including lost-in-the-middle effects \citep{liu2024lost,song2026aligning}, needle-in-a-haystack failures \citep{lee2025towards}, and inference overhead \citep{sun2025docagent}. As Figure \ref{fig:intro} illustrates, neither full-context nor local-only summarization suffices. The central challenge of streaming dialogue summarization is thus one of \emph{memory}, that is, selectively recovering the past context that the current window depends on, under a strict memory budget.

To diagnose this challenge, we examine three representative memory paradigms, \emph{recency}-based, \emph{summarization}-based, and \emph{retrieval}-based \citep{liu2024towards, wu2025incremental, li2025hierarchical}. Our analysis identifies retrieval as the strongest existing paradigm, since it preserves explicit historical evidence and selects past turns relevant to the current window. Yet, even retrieval reveals two fundamental bottlenecks under streaming settings. On the \emph{construction} side, retrieval is driven by what the window already expresses, while the memory it actually needs is defined by what the window leaves unresolved. On the \emph{utilization} side, retrieved chunks mix resolving evidence with neighboring turns, repetitions, and topic drift, lowering evidence density under a fixed memory budget.


In response, we propose \algname{} (\underline{Re}trieving \underline{M}issing \underline{E}vidence  \underline{M}emory \underline{B}y \underline{E}licited \underline{R}ecall), a context-gap-centered memory framework for streaming dialogue summarization. \algname{} operates in two stages: \emph{(i) Gap-conditioned evidence retrieval}, which elicits unresolved dependencies in the current window and issues targeted queries to retrieve historical chunks that resolve what standard retrieval overlooks; and \emph{(ii) Gap-conditioned chunk refinement}, which reduces memory noise by identifying the most relevant turn within each retrieved chunk and retaining only those turns within a fixed memory budget.
This shifts memory construction from similarity-based chunk retrieval to evidence-dense memory construction. Memory stores not history that resembles the window, but evidence that completes it. Experiments on long-context chit-chat and workplace dialogues spanning up to 160K tokens show that \algname{} improves memory recall and gap-resolution completeness over recency-, summarization-, and retrieval-based memory baselines under the same budget.

Our main contributions are summarized as: 
(1) We introduce streaming dialogue summarization as a new task with a benchmark that separately evaluates memory evidence and summary reflection; (2) We analyze representative memory paradigms and identify two bottlenecks of retrieval-based memory in streaming settings: similarity-driven construction and noisy memory utilization; (3) We propose \algname{}, a missing-evidence memory construction framework using gap-conditioned retrieval and turn-level chunk refinement to extract resolving evidence under a fixed budget; (4) We show that \algname{} improves memory recall and gap-resolution completeness over memory construction baselines, with consistent gains up to 160K tokens histories.

\section{Related Work}
\label{sec:relatedwork}
\noindent \textbf{Long Dialogue Tasks.}
Long dialogue research studies how models use information when conversations exceed a single context window. Existing work includes multi-session response generation~\citep{xu2022beyond, chen2025post, liu2025persona}, which predicts the next utterance from accumulated history, long-dialogue understanding such as QA or entity tracking over a complete conversation~\citep{kim2024dialsim, maharana2024evaluating, bai2025longbench, lee2025realtalk}, and conversational search, which rewrites prior turns into explicit retrieval queries~\citep{yoon2025ask, zhu2025convsearch}. Streaming dialogue summarization departs from these settings in two respects. No retrieval target is specified, and the model must identify what the current window leaves unresolved. The output is not a response to a question but a coherent completion of the window, rendering its content interpretable rather than answering it.

\smallskip
\noindent \textbf{Dialogue Summarization.}
Dialogue summarization initially focused on compressing dialogue-specific structures, including multi-turn dependencies, speaker roles, topic shifts, and discourse relations~\citep{tang2022confit,jia2023taxonomy,gao2023dialogue,lin2023topic,xiao2024baichuan2,yi2025survey}. 
For longer dialogues, prior work scales summarization through incremental updates of a running summary~\citep{hwang2024enhancing,ravaut2024context} or hierarchical compression of segmented chunks~\citep{zhu2020hierarchical,li2021hierarchical,zhang2021exploratory}. Query-focused and instructive dialogue summarization assumes an explicit information need is given with the input~\citep{zhong2021qmsum,wang2023instructive}. 
However, these methods are largely formulated in closed settings, where the complete dialogue is available and the output is a global summary. This assumption does not specify which past information is necessary for interpreting a particular target window, making it insufficient for streaming dialogue summarization.

\smallskip
\noindent \textbf{Long-Context Memory Construction.}
Long-context memory modeling selects or compresses past information. Summarization-based memory maintains compact histories through recursive or bounded summaries~\citep{wang2025recursively,wang2026bounded}, but compression may remove evidence needed for local contextual gaps. Retrieval-based memory fetches relevant past utterances~\citep{zhong2024memorybank,tan2025prospect,pan2025secom,li2025hello}, but high-recall retrieval can introduce irrelevant or redundant context. Intent- and goal-driven memory improves selectivity through missing slots or user-centric needs~\citep{du2026memguide,yan2026adamem}, but targets agent response generation rather than contextual evidence construction for summarization. Structured memory organizes history through timelines or self-questioning~\citep{ong2025towards,yang2026beyond}, but is optimized for general context management. These methods motivate long-history memory construction, but they do not construct evidence for resolving target-window gaps.

\smallskip
\noindent \textbf{Query Decomposition for Retrieval.} 
Retrieval-oriented query reformulation improves evidence access by rewriting underspecified or complex inputs into explicit retrieval queries~\citep{elgohary2019can,yoon2025ask,zhu2025convsearch}. Related methods also use generated queries or hypothetical documents to guide retrieval~\citep{mao2021generation,gao2023precise,wang2023query2doc}. \algname{} shares the goal of targeted retrieval, but differs in its trigger and target. It is triggered by discourse incompleteness in the current window, and its queries specify non-hypothetical evidence needs for missing prior dialogue context. This distinction matters because streaming summaries must recover referents, prior states, and rationales from earlier turns rather than infer plausible answers.

\section{Streaming Dialogue Summarization: Formulation and Evaluation}
\label{sec:streaming_dialogue_summarization}

In the absence of a formal task definition and dedicated benchmark for {streaming dialogue summarization}, we present the problem formulation (Section~\ref{sec:problem_formulation}), benchmark construction (Section~\ref{sec:benchmark}), and evaluation metrics (Section~\ref{sec:evaluation_metrics}).

\subsection{Problem Formulation}
\label{sec:problem_formulation}

We formulate \emph{streaming dialogue summarization} as a sequential task over an utterance stream. At each step $t$, the model observes a current window $\mathcal{W}_t$ of recent utterances, while all preceding utterances form the accumulated history $\mathcal{H}_t$. The goal is to generate a self-contained summary of $\mathcal{W}_t$ by leveraging history $\mathcal{H}_t$.
The summary reflects the \emph{salient utterances} within $\mathcal{W}_t$. A subset of these salient utterances is independently comprehensible. The interpretation of the remaining salient utterances strictly depends on prior context in $\mathcal{H}_t$. This dependence introduces \emph{contextual gaps}. Resolving such gaps necessitates retrieving missing evidence from $\mathcal{H}_t$. 
The history $\mathcal{H}_t$ grows unboundedly over time, rendering exhaustive access computationally impractical. Therefore, the model must construct and maintain a memory module $\mathcal{M}_t$. Instead of storing all past interactions, $\mathcal{M}_t$ extracts selectively and preserves the critical evidence necessary to resolve contextual gaps. The success of the task is determined by how effectively $\mathcal{M}_t$ provides the missing context needed to construct summary $S_t$.

\subsection{Benchmark Construction}
\label{sec:benchmark}

Streaming dialogue summarization requires utterance-level grounding over long-term dialogue histories, which existing dialogue summarization benchmarks generally lack. Without links between history-dependent utterances in $\mathcal{W}_t$ and resolving evidence in $\mathcal{H}_t$, we cannot assess whether memory is adequate or whether the summary is self-contained. We therefore construct a benchmark with explicit utterance--evidence links.

\smallskip
\noindent\textbf{Streaming Dialogue Data.}
We source 35 long dialogues from chit-chat and workplace domains across three datasets~(20 chit-chat from LoCoMo~\citep{maharana2024evaluating} and REALTALK~\citep{lee2025realtalk}; 15 workplace collaborative dialogues from EverMemBench~\citep{hu2026evermembench}). The two domains are selected because workplace dialogues sustain a single topic across sequential agendas, concentrating evidence contiguously in $\mathcal{H}_t$, whereas chit-chat dialogues shift topics frequently, dispersing it discontinuously.
Each dialogue is segmented into 1,024-token windows, each approximating five minutes of conversation, where $\mathcal{H}_t$ comprises all preceding utterances. Windows with $|\mathcal{H}_t| <$ 7K tokens or entirely self-contained salient utterances are excluded. The resulting 1,079 candidates span $\mathcal{H}_t$ lengths of 7K to 160K tokens, ensuring diversity in history length.
From these, 900 instances are sampled with domain stratification, reflecting that workplace dialogues are on average ten times longer. Detailed statistics are provided in Appendix~\ref{sec:appendix_dataset}.

\smallskip
\noindent\textbf{Gap Annotation.}
Evaluating whether a generated summary is contextually self-contained requires two references. \emph{Salient utterances} define the current-window content to preserve, and \emph{contextual gaps} specify the missing historical information needed to make history-dependent utterances self-contained. Together, they serve as ground truth for memory adequacy and summary quality.

A contextual gap is annotated only when incompleteness stems from missing prior context. Following completeness-related error types in prior dialogue summarization studies~\citep{liu2021coreference, tang2022confit, zhu2023annotating, kirstein2024s}, we define three history-resolvable gap types. (i)~Referential gaps denote missing antecedents of pronouns, mentions, or discourse references. (ii)~Attribute gaps refer to missing states, preferences, or entity properties. (iii)~Relational gaps encompass missing causal, logical, or discourse-level relations. Each history-dependent salient utterance receives one representative gap, resolved by one or more evidence utterances from $\mathcal{H}$. Appendix~\ref{sec:appendix_taxonomy} provides definitions and examples.

Salient utterances and contextual gaps are constructed through a five-stage LLM-assisted annotation pipeline, detailed in Appendix~\ref{sec:appendix_annotation}. Each stage employs three independent LLMs from distinct model families. Outputs are retained only by majority agreement~\citep{kim2024prometheus,lee2024unisumeval,thakur2025judging}. This design reduces single-family dependence and yields a reproducible reference standard for comparing memory construction methods. Each retained gap records the target salient utterance, gap type, resolving evidence utterances from history, and resolution statement. The final benchmark instance pairs the current window and accumulated history with the reference salient utterances and contextual gaps.

\subsection{Evaluation Metrics}
\label{sec:evaluation_metrics}

Summary quality alone is insufficient to evaluate streaming summarization, as it cannot distinguish memory-grounded gap resolution from confabulated context. We therefore evaluate two axes, memory quality via \emph{memory recall}, and summary quality measured by \emph{faithfulness}, \emph{conciseness}, and \emph{completeness}~\citep{song2024finesure, min2025towards}. To isolate the effect of memory on summarization, completeness is decomposed into \emph{window} completeness and \emph{gap-resolution} completeness. All metrics are detailed in Appendix~\ref{sec:appendix_metrics}.

\smallskip
\noindent\textbf{Memory Recall.}
Memory quality is assessed independently of the generated summary, as evidence recovery and summary generation are 
distinct failure modes. For each annotated contextual gap, we assign a binary label indicating whether the memory satisfies the corresponding resolution statement. Memory recall is the proportion of positively labeled gaps, measuring how completely the memory module preserves the historical basis for self-contained summarization.

\smallskip
\noindent\textbf{Window Completeness.}
Independent of memory, a summary must also cover the salient content observable in the current window. For each salient utterance, we check whether its content appears in the summary. Window completeness is the fraction of salient utterances covered, measuring how well the summary reflects the current window, regardless of memory-related failures.

\smallskip
\noindent\textbf{Gap-Resolution Completeness.}
Recovered evidence is only meaningful if subsequently reflected in the summary. Among gaps resolved by memory, we check whether the resolution is expressed in the summary. Gap-resolution completeness is the fraction of memory-resolved gaps reflected in the summary, measuring how well the model utilizes recovered evidence in generation.

\section{Methodology: \algname{}}
\label{sec:methodology}
The quality of $\mathcal{M}_t$ constrains whether the summarizer receives gap-resolving evidence. Summarization-based memory can dilute such evidence through global compression, while retrieval with window-level queries preserves detail but favors window-overlapping history over gap-resolving evidence. \algname{} addresses this mismatch by constructing memory around contextual gaps, retrieving missing evidence with gap-specific queries, and refining retrieved chunks into a compact $\mathcal{M}_t$ under the memory budget.

\subsection{Gap-Conditioned Evidence Retrieval}
\label{sec:gap_conditioned_evidence_retrieval}
Retrieval in streaming dialogue summarization suffers from a \emph{history-window mismatch}. A query derived from $\mathcal{W}_t$ surfaces visible window content and retrieves history that repeats or paraphrases it. Yet the needed evidence is the presupposed referent, state, rationale, or causal link. \algname{} addresses this mismatch by conditioning retrieval on detected contextual gaps rather than the full $\mathcal{W}_t$.

\smallskip
\noindent\textbf{Gap Detection and Query Construction.}
Gap-conditioned retrieval proceeds in two stages. The first identifies which utterances are both summary-worthy and gap-bearing. The second constructs an evidence query for each identified gap without hypothesizing its resolution.
An utterance is \emph{summary-worthy} when it conveys information that should be preserved in a self-contained summary, such as an action, decision, plan, or status update. It is \emph{gap-bearing} when its interpretation depends on prior dialogue, such as an absent referent, rationale, prior state, or causal dependency. These criteria define the retrieval trigger at inference time. They are applied only to $\mathcal{W}_t$, without benchmark annotations, resolving evidence, or gap-type labels.

Detecting such utterances requires discourse-level reasoning beyond lexical overlap, since the missing evidence may not share surface form with any token in $\mathcal{W}_t$~\citep{elgohary2019can, du2026memguide}.
\algname{} implements this detection with a compact LLM to keep per-window inference cost tractable\footnote{We use Qwen3.5-4B~\citep{qwen3.5} as the default for gap detection; ablations with Qwen3.5-9B and Gemma-4-E2B-it~\citep{gemma4} show no consistent gain over the 4B scale~(Appendix~\ref{sec:appendix_gep_detection_model_ablation}).}.
For each gap-bearing utterance, \algname{} converts the unresolved dependency into an evidence-seeking query. The query identifies the gap-bearing utterance, states why the utterance is not self-contained, specifies the type of evidence required for resolution, and supplies lexical anchors drawn from $\mathcal{W}_t$. 
This avoids anchoring retrieval to a hypothetical answer that may not exist in $\mathcal{H}_t$, reducing retrieval drift. 
The prompt used for gap detection is shown in Figure~\ref{fig:prompt_remember}.

\smallskip
\noindent\textbf{Candidate Chunk Construction.}
Candidate construction follows three design choices. First, $\mathcal{H}_t$ is segmented into 128-token chunks with 32-token overlap. This granularity keeps each chunk small enough to isolate specific evidence while preserving sufficient turn context for retrieval. Second, each gap query is issued to both a sparse and a dense retriever, targeting complementary aspects of relevance. 
Sparse retrieval uses BM25~\citep{robertson2009probabilistic} to enforce lexical overlap with the query anchors. Dense retrieval uses Qwen3-Embedding-0.6B~\citep{zhang2025qwen3}, a model that supports instruction-aware embeddings and is therefore well-suited to the structured, multi-part gap queries\footnote{Each retriever returns the top 30 chunks per gap, providing a bounded candidate pool that is broad enough for evidence recall without approximating full-history access.}. 
Third, the two ranked lists are fused with reciprocal rank fusion, which aggregates ranks without cross-system score calibration. For each gap query $g$, the fused score for a chunk $c$ is computed as $s_g(c) = \frac{1}{60 + r^{\mathrm{sparse}}(c)} + \frac{1}{60 + r^{\mathrm{dense}}(c)},$ where $r^{\mathrm{sparse}}(c)$ and $r^{\mathrm{dense}}(c)$ denote the ranks of $c$ under the respective retrievers, following the standard setting for reciprocal rank fusion~\citep{cormack2009reciprocal}. 
The top-$K$ chunks per gap are retained as $\mathcal{C}_{g}$, where $K$ is determined by the token budget divided by the chunk size.

\subsection{Gap-Conditioned Chunk Refinement}
\label{sec:gap_aware_evidence_distillation}

Retrieved chunks introduce a second failure mode, \emph{chunk-evidence mismatch}. A 128-token chunk provides sufficient context for retrieval. Yet it inevitably carries gap-irrelevant content that occupies memory space without contributing to gap resolution~\citep{pan2025secom, song2026aligning}. \algname{} therefore refines $\mathcal{C}_g$ through two steps, gap-conditioned turn extraction and gap-balanced evidence accumulation.

\smallskip
\noindent \textbf{Gap-Conditioned Turn Extraction.}
Scoring at the chunk level dilutes the gap-resolution signal, since the required clue is often concentrated in a single utterance.
\algname{} decomposes each chunk $c_i \in \mathcal{C}_g$ into constituent utterances.
Each utterance and the gap query $g$ are embedded with the same Qwen3-Embedding-0.6B encoder used in the retrieval step. 
\algname{} then scores each utterance as $\cos(e(u),\,e(g))$. Since $g$ already encodes the specific evidence need that triggered retrieval, this cosine score serves as a sufficient gap-conditioned relevance signal without additional supervision.
The utterances within each $c_i$ are thus re-ranked per gap, pushing gap-irrelevant content down. The complete turn containing each high-scoring utterance is then extracted, including its speaker information, as a turn-level evidence unit. Thus, $\mathcal{C}_g$ is converted into $\mathcal{T}_g$, a set of turn-level evidence units ordered by gap relevance.

\smallskip
\noindent \textbf{Gap-Balanced Evidence Accumulation.}
Turns are accumulated into $\mathcal{M}_t$ in rounds, where round $d$ appends the $d$-th highest-scoring turn from each $\mathcal{T}_g$ if not already present, until the prespecified memory budget $B$ is reached.
This yields $\mathcal{M}_t = \{\tau_{g,d} \mid g \in \mathcal{G}_t,\ d = 1, \ldots, D\}$,
where $D$ is the largest round completed before $\mathcal{M}_t$ exceeds the token budget $B$.
This round-robin allocation ensures that no single gap monopolizes $\mathcal{M}_t$, so each unresolved dependency in $\mathcal{W}_t$ retains representation. 
The resulting $\mathcal{M}_t$ is thus not a general compression of $\mathcal{H}_t$ but a gap-aligned collection of turn-level evidence, enabling the summary generator to produce a contextually self-contained summary of $\mathcal{W}_t$ without access to the full $\mathcal{H}_t$.

\subsection{Integration with Summary Generation}
\algname{} is designed as a gap-resolving memory module. The constructed memory $\mathcal{M}_t$ can be integrated with any standard summary generator. In our experiments, generation is held fixed across memory strategies. The generator receives only $\mathcal{M}_t$ and the current window $\mathcal{W}_t$ as input. The prompt for summary generation can be found in Figure~\ref{fig:prompt_summary_generation_w_memory}.

\section{Evaluation}
\label{sec:experiments}
In this section, we evaluate whether \algname{} constructs gap-resolving memory and improves streaming dialogue summaries under a fixed memory budget.
Section~\ref{sec:experimental_setup} introduces metrics and baselines. Sections~\ref{sec:memory_quality}--\ref{sec:summary_quality} evaluate memory recall and summary quality against memory construction baselines. Sections~\ref{sec:gap_analysis}--\ref{sec:ablation} further analyze recall breakdowns by gap type, component-level contributions and computational cost. 
\subsection{Experimental Setup}
\label{sec:experimental_setup}

The experiment isolates the effect of memory construction by holding the summarizer fixed across all methods. Each method receives the same current window $\mathcal{W}$ and constructs $\mathcal{M}$ from prior dialogue. 

\smallskip
\noindent \textbf{Metrics.}
Memory quality is measured by \emph{memory recall} and summary quality along four dimensions (\emph{window completeness}, \emph{gap-resolution completeness}, \emph{faithfulness}, and \emph{conciseness}), as defined in Section~\ref{sec:evaluation_metrics}. We additionally report \emph{composite score}, the arithmetic mean across all four dimensions, as an overall measure of summary quality.

All metrics are scored by Qwen3.6-27B~(temperature 0.0) with rubric-guided prompting. Rubric-guided LLM evaluation has been shown to achieve human-level correlation across NLG tasks~\citep{zheng2023judging, kim2024prometheus}, with stronger models further improving agreement with human annotators~\citep{thakur2025judging} and consistently strong alignment on faithfulness and completeness~\citep{song2024finesure, lee2024unisumeval}. Scoring criteria and prompts are provided in 
Figures~\ref{fig:prompt_memory_recall}-\ref{fig:prompt_fact}.


\smallskip
\noindent \textbf{Baselines.}
We compare \algname{} against two groups. The first group consists of \emph{diagnostic references} that control memory
availability and quality. The second group consists of \emph{memory construction baselines} that build memory under the same capacity constraint.

\textbf{(i) Diagnostic references} isolate how summary quality changes as a function of memory availability and organization, independently of any construction method. \texttt{Ideal Memory} supplies the reference gap-resolving evidence as memory, serving as an upper-bound diagnostic for evidence availability. \texttt{No Memory} summarizes $\mathcal{W}_t$ alone, measuring the summary quality achievable without any historical context. \texttt{Full Memory} supplies the entire $\mathcal{H}_t$ as memory, revealing whether the bottleneck lies in evidence availability or in how evidence is organized for generation. When $\mathcal{H}$ exceeds the model context limit, the oldest turns are truncated first.

\textbf{(ii) Memory construction baselines} represent three representative strategies for converting $\mathcal{H}_t$ into a bounded memory. \texttt{Recency-based memory} (\texttt{Recent Memory}) retains the most recent turns immediately preceding $\mathcal{W}_t$, under the rationale that temporally proximate turns are most likely to resolve current gaps.  \texttt{Summarization-based Memory} (\texttt{Inc.\,Summary}, \texttt{Hier.\,Summary}) instead compresses the entire $\mathcal{H}_t$ into a fixed-length memory, covering the full history at the cost of abstractive compression. 
\texttt{Inc.\,Summary} updates a running summary at each window, while \texttt{Hier.\,Summary} aggregates segment-level summaries bottom-up (prompt in Figure~\ref{fig:prompt_summ_based}). The compression operates over 4,096-token segments with a 1,024-token overlap, applied recursively.
\texttt{Retrieval-based Memory} (\texttt{Sparse}, \texttt{Dense}, \texttt{Hybrid}) retrieves chunks from $\mathcal{H}_t$ using $\mathcal{W}_t$ as a single composite query via lexical, semantic, and hybrid matching, respectively, as window-aware retrieval better captures the evidence signal than compression. Sparse retrieval uses BM25~\citep{robertson2009probabilistic}; dense retrieval uses Qwen3-Embedding-0.6B~\citep{zhang2025qwen3} and packs the top-8 chunks to ensure under the memory budget; hybrid retrieval packs the top-4 chunks from each ranking.
 
\noindent\textbf{Implementation Details.}
Both the memory capacity and window size are fixed at 1,024 tokens. Under this setting, \algname{} retains top-8 retrieved chunks, giving a retrieval budget of 8 chunks per query. This ensures that memory does not dominate the generation context while providing sufficient capacity for historical evidence. All memory-based methods use the same summary-generation instruction (Figure~\ref{fig:prompt_summary_generation_w_memory}). \texttt{No Memory}, which requires no memory context, uses a separate prompt (Figure~\ref{fig:prompt_summary_generation}). Summary evaluation results are averaged across three summarizers~(Qwen3.5-4B, Qwen3.5-9B, and Gemma-4-E2B-it at temperature 0.7). Further details are in Appendix~\ref{sec:appendix_model_implementation}.

\begin{table}[t!]
\scriptsize
    \centering

\setlength{\tabcolsep}{5pt}
\begin{tabular}{L{3cm}X{2.0cm}X{2.0cm}X{2.0cm}X{2.0cm}}
\toprule
{Memory Type} & Memory Recall & Referential Gap & Attribute Gap & Relational Gap \\
\midrule
\rowcolor{oraclerow}  Ideal Memory & 1.0000 & 0.3251 & 0.5421 & 0.1328 \\
 \rowcolor{oraclerow} Full Memory & 0.8445 & 0.2410 & 0.4252 & 0.0935 \\
\midrule
Recent Memory & 0.4126 & 0.1756 & 0.2130 & 0.0441 \\
\midrule
Inc. Summary & 0.3581 & 0.1355 & 0.1983 & 0.0340 \\
 Hier. Summary & 0.3212 & 0.1095 & 0.2196 & 0.0214 \\
\midrule
Sparse& 0.5130 & 0.2029 & 0.2617 & 0.0621 \\
Dense & 0.4897  & 0.1969 & 0.2490 & 0.0561 \\
Hybrid & 0.5412 & 0.2183 & 0.2777 & 0.0614 \\
\midrule
\rowcolor{proposedrow} \algname{} & 0.6984 & 0.2457 & 0.3558 & 0.1008 \\
\bottomrule
\end{tabular}
\caption{Memory recall and its breakdown by gap type, where each type-level score is the fraction of gaps included in $\mathcal{M}_t$ out of all annotated gaps of that type. Domain-wise results are in Appendix~\ref{sec:appendix_mr_domain_wise}.}
\label{tab:memory-evaluation}
\end{table}

\subsection{Gap-Resolving Evidence Recovery}
\label{sec:memory_quality} 

Table~\ref{tab:memory-evaluation} reveals \textbf{memory recall depends on how evidence is selected, not on how much history is accessed.} \algname{} achieves a memory recall of 0.6984, outperforming the best memory construction baseline, \texttt{Hybrid}, by 0.157. \texttt{Inc.\,Summary} and \texttt{Hier.\,Summary} access the full history yet fall below even the \texttt{Recent\,Memory}. This indicates that indiscriminate compression actively discards task-relevant evidence. \texttt{Sparse}, \texttt{Dense}, and \texttt{Hybrid} recover more evidence but remain bounded by surface similarity to the current window, leaving implicitly connected evidence unrecovered regardless of retrieval budget.

\begin{table*}[t!]
\tiny
    \centering

\setlength{\tabcolsep}{0.7pt}
\begin{tabular}{L{1.8cm}X{0.56cm}X{1.05cm}X{0.56cm}@{\hspace{3pt}}X{0.56cm}X{1.05cm}X{0.56cm}@{\hspace{3pt}}X{0.56cm}X{1.05cm}X{0.56cm}@{\hspace{3pt}}X{0.56cm}X{1.05cm}X{0.56cm}@{\hspace{3pt}}X{0.56cm}X{1.05cm}X{0.56cm}}
\toprule
 \multirow[c]{2}{*}{Memory Type} & \multicolumn{3}{c}{Win-Comp} & \multicolumn{3}{c}{Gap-Comp} & \multicolumn{3}{c}{Conc} & \multicolumn{3}{c}{Faith} & \multicolumn{3}{c}{Composite} \\
\cmidrule(lr){2-4} \cmidrule(l){5-7} \cmidrule(l){8-10} \cmidrule(l){11-13} \cmidrule(l){14-16}
& -32K & \mbox{32K-64K} & 64K- & -32K & \mbox{32K-64K} & 64K- & -32K & \mbox{32K-64K} & 64K- & -32K & \mbox{32K-64K} & 64K- & -32K & \mbox{32K-64K} & 64K- \\
\midrule
\rowcolor{oraclerow} Ideal Memory & 0.73 & 0.71 & 0.70 & 0.73 & 0.76 & 0.77 & 0.65 & 0.71 & 0.73 & 0.88 & 0.88 & 0.90 & 0.75 & 0.77 & 0.77 \\
\rowcolor{oraclerow} No Memory & 0.70 & 0.65 & 0.65 & N/A & N/A & N/A & 0.61 & 0.66 & 0.68 & 0.92 & 0.91 & 0.91 & 0.55 & 0.56 & 0.56 \\
\rowcolor{oraclerow} Full Memory & 0.39 & 0.21 & 0.13 & 0.45 & 0.42 & 0.24 & 0.31 & 0.22 & 0.16 & 0.76 & 0.80 & 0.80 & 0.48 & 0.41 & 0.33 \\
\midrule
Recent Memory & 0.68 & 0.68 & 0.67 & 0.21 & 0.24 & 0.40 & 0.52 & 0.60 & 0.63 & 0.88 & 0.89 & 0.90 & 0.57 & 0.60 & 0.65 \\
\midrule
Inc. Summary & 0.72 & 0.73 & 0.72 & 0.27 & 0.17 & 0.17 & 0.54 & 0.63 & 0.65 & 0.74 & 0.74 & 0.75 & 0.57 & 0.57 & 0.57 \\
Hier. Summary & 0.74 & 0.73 & 0.71 & 0.30 & 0.17 & 0.05 & 0.55 & 0.61 & 0.62 & 0.73 & 0.71 & 0.72 & 0.58 & 0.56 & 0.52 \\
\midrule
Sparse & 0.69 & 0.69 & 0.70 & 0.28 & 0.31 & 0.43 & 0.54 & 0.63 & 0.67 & 0.87 & 0.88 & 0.89 & 0.60 & 0.63 & 0.67 \\
Dense & 0.70 & 0.66 & 0.67 & 0.32 & 0.32 & 0.36 & 0.56 & 0.62 & 0.67 & 0.87 & 0.87 & 0.88 & 0.62 & 0.62 & 0.64 \\
Hybrid & 0.70 & 0.68 & 0.69 & 0.31 & 0.31 & 0.45 & 0.55 & 0.63 & 0.67 & 0.88 & 0.90 & 0.90 & 0.61 & 0.63 & 0.68 \\
\midrule
\rowcolor{proposedrow} \algname{} & 0.73 & 0.72 & 0.72 & 0.41 & 0.48 & 0.47 & 0.58 & 0.66 & 0.68 & 0.88 & 0.88 & 0.89 & 0.65 & 0.68 & 0.69 \\
\bottomrule
\end{tabular}
\caption{Summary quality across dialogue length bins (-32K, 32K--64K, 64K-). Win-Comp, Gap-Comp, Conc, and Faith denote window completeness, gap-resolution completeness, conciseness, and faithfulness, respectively. For \texttt{No Memory}, the unavailable Gap-Comp is counted as 0 as no historical evidence is supplied. Scores are averaged over three summarizers, and higher scores indicate better performance. Model-wise results are in Appendix~\ref{sec:appendix_summarizer_wise}}
\label{tab:summary-length}
\end{table*}

\subsection{Impact on Streaming Summaries}
\label{sec:summary_quality}
Table~\ref{tab:summary-length} reports summary quality across dialogue length bins.
\algname{} achieves the best composite score among memory construction baselines in every length bin. This advantage is driven mainly by gap-resolution completeness, where \algname{} improves over \texttt{Hybrid} by up to 0.17, while maintaining strong window completeness, conciseness, and faithfulness. This pattern indicates that the summary gains come from resolving missing contextual dependencies without sacrificing current window coverage or reliability.
In contrast, \texttt{Full Memory} exposes substantially more history, but achieves lower window completeness and conciseness than \texttt{Retrieval-based Memories}. The diagnostic references show that \textbf{the central challenge is not evidence availability alone, but organizing historical evidence into a form that the summarizer can reliably use.}


\begin{table}[t!]
\scriptsize
    \centering

\setlength{\tabcolsep}{3pt}
\begin{tabular}{L{2.5cm}X{1.5cm}X{1.5cm}X{1.5cm}X{1.5cm}X{1.5cm}X{1.5cm}}
\toprule
Memory Type & \multicolumn{2}{c}{Referential Gap} & \multicolumn{2}{c}{Attribute Gap} & \multicolumn{2}{c}{Relational Gap} \\
\cmidrule(lr){2-3} \cmidrule(l){4-5} \cmidrule(l){6-7}
 &  Memory & Summary & Memory & Summary & Memory & Summary \\
\midrule
\rowcolor{oraclerow} Ideal Memory & 100.0 & 80.34 & 100.0 & 73.73 & 100.0 & 72.96 \\
\rowcolor{oraclerow} Full Memory & 78.80 & 35.73 & 83.79 & 37.11 & 74.03 & 21.45 \\
\midrule
Recent Memory &51.15 & 37.31 & 39.10 & 26.90 & 33.79 & 22.71 \\
\midrule
Inc. Summary & 42.24 & 26.96 & 35.98 & 21.01 & 25.14 & 12.89 \\
Hier. Summary &33.24 & 21.21 & 36.38 & 19.34 & 16.02 & 7.00 \\
\midrule
Sparse & 60.33 & 42.77 & 48.2 & 31.96 & 45.95 & 28.58 \\
Dense & 58.61 & 43.82 & 44.99 & 30.30 & 42.17 & 26.00 \\
Hybrid & 64.32 & 46.34 & 51.29 & 34.46 & 46.13 & 25.68 \\
\midrule
\rowcolor{proposedrow} \algname{} &  75.26 & 57.15 & 66.15 & 46.78 & 76.89 & 52.21 \\
\bottomrule
\end{tabular}
\caption{Percentage of annotated gaps per type whose resolving evidence is included in memory and reflected in the generated summary.}
\label{tab:gap-type-memory-summary-percent}
\end{table}

\begin{table}[t!]
    \scriptsize
    \centering
    \setlength{\tabcolsep}{2pt}
    \begin{tabular}{L{4.5cm}X{2.0cm}X{1.5cm}X{1.5cm}X{1.0cm}X{1.0cm}}
    \toprule
    Variations & Memory Recall & Gap-Comp & Win-Comp & Conc & Faith \\
    \midrule
         (i) w/o Gap Conditioned Retrieval & 0.60 & 0.43 & 0.63 & 0.62 &0.82 \\
         (ii) w/o Chunk Refinement & 0.66 & 0.46 & 0.64 & 0.66 &0.82 \\
         \midrule
         \algname{} &  0.70 & 0.50 & 0.68 & 0.66 &0.81 \\
    \bottomrule
    \end{tabular}
    \caption{Ablation results using Qwen3.5-4B as gap detection model and summarizer.}
    
    \label{tab:ablation}

\end{table}

\subsection{Gap-Level Evidence Use}

\label{sec:gap_analysis}
Table~\ref{tab:gap-type-memory-summary-percent} separates evidence inclusion in memory from evidence reflection in the summary. \algname{} improves most in this regime by reasoning beyond lexical overlap, raising memory inclusion and summary coverage for relational gaps over memory construction baselines. \texttt{Full Memory} nonetheless buries gap-resolving evidence among irrelevant context across all gap types, producing a needle-in-a-haystack failure~\citep{lee2025towards}. Even \texttt{Ideal Memory} does not guarantee full evidence reflection, showing that generation imposes a separate ceiling. \textbf{The persistent gap between memory inclusion and summary use suggests that full resolution requires co-design of memory construction and summary generation.}


\subsection{Ablation Study}
\label{sec:ablation}
Table~\ref{tab:ablation} isolates the contribution of each component in \algname{}. \textit{(i) w/o Gap-Conditioned Retrieval} retrieves with the current window as a single query and keeps the same memory budget; \textit{(ii) w/o Chunk Refinement} stores retrieved chunks without refinement. 
Removing gap-conditioned retrieval causes the larger drop. Window-level retrieval tends to recover history overlapping with $\mathcal{W}_t$, rather than evidence that resolves its implicit dependencies. 
Removing chunk refinement also degrades performance, indicating that raw chunks contain gap-irrelevant context that consumes the fixed memory budget. These results show that \algname{} improves memory construction by targeting missing evidence needs and densifying retrieved evidence.

\subsection{Latency--Memory Recall Trade-off}
\label{sec:latency_tradeoff}
\begin{figure}[t!]
    \centering
    \includegraphics[width=0.60\linewidth]{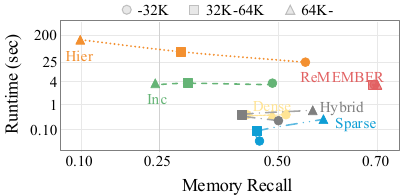}

\caption{Memory recall versus memory build runtime (log scale, in seconds) across three dialogue length bins.}
    \label{fig:latency_tradeoff}
\end{figure}

Figure~\ref{fig:latency_tradeoff} compares memory recall and memory-construction runtime across three dialogue-length bins.\footnote{Single H200 GPU; one CPU thread.} 
\textbf{Gap-conditioned construction achieves a favorable balance between recall and runtime as dialogue length grows}.
\algname{} incurs a construction latency of approximately four seconds across all tested dialogue-length bins, yet consistently achieves the highest memory recall. 
Retrieval-based methods~(\texttt{Sparse}, \texttt{Dense}, \texttt{Hybrid}) run in under one second but recover less gap-resolving evidence, whereas \texttt{Hier.\,Summary}~(Hier) incurs the highest runtime while yielding the lowest recall, indicating that repeated compression is both costly and lossy under a fixed memory budget.

\section{Conclusion}
\label{sec:conclusion}
We present {streaming dialogue summarization} as a new task with a benchmark and evaluation metrics. 
Our analysis shows that existing memory paradigms retrieve evidence aligned with what the current window states, not what it leaves implicit. 
To address this mismatch, \algname{} constructs memory around detected contextual gaps rather than surface similarity. 
Under the same budget, \algname{} improves memory recall and gap-resolution completeness across histories up to 160K tokens. 
Even with reference gap-resolving evidence, a persistent gap remains between memory recall and summary use, indicating that summary generation independently limits gap resolution.

\bibliographystyle{assets/plainnat}
\bibliography{colm2026_conference}
\clearpage
\appendix

\begin{table*}[t!]
\scriptsize
\centering
\setlength{\tabcolsep}{1pt}
\begin{tabular}{L{2.0cm}X{2.0cm}X{1.0cm}X{2.5cm}X{1.0cm}X{1.0cm}X{1.0cm}X{1.0cm}X{1.0cm}}
\toprule
Domain & Source &\# Samples & Avg. Dialogue Length & Ref. Gap  & Att. Gap  & Rel. Gap  &  Avg. Sal. & Avg. Gap \\
\midrule
Chit-Chat & LoCoMo, RealTalk21 & 300  &  20,446 (8,024-43608) &  205 & 188 & 71 & 4.70 & 1.55 \\
Workplace & EverMemBench & 600 &  80,574 (8,024-159,576) &   282 & 624 & 128 & 9.49 & 1.73 \\
\midrule
All & -- & 900 & 60,532 (8,024-159,576) & 487 & 812 & 199 & 7.90 & 1.67 \\
\bottomrule
\end{tabular}
\caption{Benchmark statistics by domain. Dialogue length reports the average number of tokens in each benchmark instance, including both the accumulated history and the current window, with the minimum--maximum range in parentheses. Referential, Attribute, and Relational gaps denote the number of annotated contextual gaps by type. Ref. Gap, Att.Gap, Rel. Gap, Avg. Sal. and Avg. Gap denote Referential Gap, Attribute Gap, Relational Gap, the average number of salient utterances and contextual gaps per instance, respectively.}
\label{tab:dataset_statistics}
\end{table*}

\label{sec:appendix}

\section{Benchmark Construction Detail}
\subsection{Benchmark Statistics}
\label{sec:appendix_dataset}
This section provides additional details on the benchmark dataset described in Section~\ref{sec:benchmark}.
Table~\ref{tab:dataset_statistics} reports the source datasets and benchmark statistics by domain. 
The benchmark contains 900 window-level instances sampled from 35 source dialogues~(10 LoCoMo~\citep{maharana2024evaluating}, 10 RealTalk21~\citep{lee2025realtalk}, and 15 EverMemBench~\citep{hu2026evermembench}). 
Chit-chat instances are drawn from LoCoMo and RealTalk21, while workplace instances are drawn from EverMemBench. 
Dialogue length is measured at the benchmark-instance level as $|\mathcal{H}_t| + |\mathcal{W}_t|$, with the current window fixed to 1,024 tokens, rather than as the length of the source dialogue. 
Table~\ref{tab:model_implementation} lists the original dataset repositories used to obtain the raw data before preprocessing.
  
\subsection{Gap Taxonomy}
\label{sec:appendix_taxonomy}
This section provides detailed definitions and examples for the contextual gap types introduced in Section~\ref{sec:benchmark}. 
Building on prior work on faithfulness error types in dialogue summarization~\citep{liu2021coreference, tang2022confit, zhu2023annotating,kirstein2024s}, we reinterpret the underlying discourse phenomena as pre-generation contextual losses in streaming dialogue summarization. Prior work analyzes faithfulness failures after generation, including missing information, incorrect entity or circumstance details, and unsupported links in generated summaries. 
Our setting instead focuses on the corresponding omissions before generation. The current window may omit the antecedent needed to resolve a reference, the attribute needed to specify an entity or event, or the relation needed to understand why an utterance follows. 
We therefore annotate history-dependent incompleteness as a contextual gap that can be resolved by evidence from $\mathcal{H}_t$, rather than as an error introduced by the summarizer.

We organize contextual gaps into three history-resolvable dependency types according to what is omitted from the current window. Prior context may supply a missing referent, a missing attribute, or a missing relation. These roles correspond to referential, attribute, and relational gaps, respectively. Table~\ref{tab:contextual_gap_examples} provides annotated examples across all three types.

\noindent $\bullet$ \textbf{Referential gaps} arise when an utterance contains an expression whose antecedent or referent is absent from $\mathcal{W}_t$. Such expressions include pronouns, shorthand references, vague noun phrases, demonstratives, and discourse references. These gaps ask which entity, event, decision, request, or prior statement the utterance refers to.

\noindent $\bullet$ \textbf{Attribute gaps} arise when the relevant entity or event is identifiable, but a necessary property, parameter, state, preference, constraint, scope, role, assignment, or specification is absent from $\mathcal{W}_t$. These gaps ask what attribute or specification is needed to make the utterance self-contained.

\noindent $\bullet$ \textbf{Relational gaps} arise when the current utterance depends on an omitted causal, temporal, logical, or discourse relation. The missing relation may involve prior reasoning, motivation, agreement, comparison, condition, or constraint. These gaps ask why the utterance follows, how it is justified, or how it connects to prior dialogue.

If a salient utterance contains multiple distinct missing dependencies, they are identified separately during annotation. For evaluation, each history-dependent salient utterance is associated with one representative gap and one or more pieces of evidence from $\mathcal{H}_t$ that explicitly resolve the missing referent, attribute, or relation.

\begin{table*}[t!]
  \scriptsize
  \centering
  \renewcommand{\arraystretch}{1.5}
  \setlength{\tabcolsep}{3.5pt}
  \begin{tabular}{L{1.4cm}L{2.5cm}L{3.2cm}L{2.5cm}L{3.0cm}}
  \toprule
  Gap Type & Gap Utterance & Evidence & Gap Reason & Resolution \\
  \midrule
  Referential Gap
  & During testing, be sure to have the other party simulate a few requests with incorrect signatures and non-standard data formats.
  & We are officially starting a new development task today to implement a Webhook interface to receive order notifications generated by external partner channels.
  & The phrase ``the other party'' cannot be stably resolved from the current utterance alone.
  & The evidence identifies the relevant counterpart as the external partner channels that generate order notifications. \\
  \midrule
  Attribute Gap
  & I plan to reuse the distributed lock solution used when creating orders previously to ensure the idempotency of callback processing.
  & I'm focusing on testing the idempotency logic for payment callbacks, especially the solution you emphasized, which uses Redis distributed locks combined with a unique index for the database payment transaction
  number.
  & The target mentions a prior distributed lock solution, but omits its technical specification.
  & The evidence specifies that the solution combines Redis distributed locks with a unique database index. \\
  \midrule
  Relational Gap
  & @All members Synchronizing the final decision on the selection logic of emission factors.
  & Technically, @Ruiqing Jiang, when designing the algorithm, you need to reserve an interface to prioritize user-defined factors, and if none exist, then fall back to our built-in database.
  & The target announces a final decision, but the decision logic itself is not stated.
  & The evidence supplies the relation that defines the decision, namely prioritizing user-defined factors before falling back to the built-in database. \\
  \bottomrule
  \end{tabular}
  \caption{Examples of contextual gap types. The gap utterance is from the current window $\mathcal{W}_t$, while the evidence is from the prior history $\mathcal{H}_t$.}
  \label{tab:contextual_gap_examples}
\end{table*}

\subsection{Gap Annotation Pipeline Details} 
\label{sec:appendix_annotation}
We provide full implementation details for the five-stage annotation pipeline introduced in Section~\ref{sec:benchmark}. In all stages, a candidate is retained only when at least two of the three LLMs agree; disagreements are resolved by the majority label without further adjudication. 

\noindent\textbf{(i) Salient Utterance Identification.} We select utterances from $\mathcal{W}_t$ that contain a concrete decision, request, action item, or status update. Simple acknowledgments, greetings, praise, and emotional reactions are excluded. When consecutive utterances form a single coherent update or decision, they are merged into one entry. 

\noindent\textbf{(ii) Contextual Dependence Judgment.} For each salient utterance, we assess whether it meets the standard of self-contained summarizability, defined as the ability to be summarized clearly and stably from $\mathcal{W}_t$ alone without recourse to earlier dialogue. Each utterance is examined for unresolvable pronouns, vague noun phrases, underspecified attributes, and unstated reasoning that would prevent independent summarization. A gap is assigned only when the missing information is not recoverable from nearby utterances within $\mathcal{W}_t$; additional detail that would merely enrich the summary does not qualify. Each gap is recorded under one of three types following the taxonomy in Appendix~\ref{sec:appendix_taxonomy}: referential, attribute, or relational. An utterance is marked as history-dependent only when at least two LLMs agree on the same gap type. Each history-dependent utterance is assigned at most one representative gap, selected as the dependency whose absence most severely impairs summarizability.

Stages~(i) and~(ii) are implemented jointly in a single prompt across three diverse LLMs, Qwen3.6-27B~\citep{qwen3.6-27b}, Gemma-4-31B-it~\citep{gemma4}, and GLM-4.7-Flash~\citep{5team2025glm45agenticreasoningcoding}, each run at temperature 0.0; the prompt is provided in Figure~\ref{fig:prompt_salient}. 

\noindent\textbf{(iii) Candidate History Filtering.} We screen all chunks in $\mathcal{H}_t$ for relevance to $\mathcal{W}_t$ and retain only those warranting further examination in Stage~(iv). Each chunk receives a binary label independently. A chunk is labeled relevant when it contains prior content that is necessary or directly useful for understanding or summarizing $\mathcal{W}_t$, such as a decision or plan explicitly continued in the window, information needed to resolve references, or earlier steps of an ongoing task. Topical similarity alone does not qualify; a chunk is discarded when it is only loosely related or when connecting it to the window requires inference beyond what the dialogue supports. The three LLMs are drawn from distinct model families, Qwen3.6-27B, Gemma-4-31B-it, and GLM-4.7-Flash, and each is run at temperature 0.0. The prompt is provided in Figure~\ref{fig:prompt_relevant_chunk_filtering}.

\noindent\textbf{(iv) Evidence and Resolution Annotation.} For each history-dependent utterance, we audit every sentence within the retained candidate chunks to identify those that explicitly supply the missing information. Each sentence is evaluated independently through a two-step protocol. The judge first writes a reasoning trace specifying what missing content the sentence provides or fails to provide with respect to the identified gap, then assigns a binary label of \textit{resolve} or \textit{not\_resolve} based solely on that trace. Topical overlap, shared entities, discourse proximity, and broad background context do not constitute resolution; a sentence qualifies only when it explicitly provides the exact referent, attribute, prior decision, or commitment that the target utterance presupposes. The reasoning traces of all positively labeled sentences are subsequently aggregated into a single resolution statement for the gap. Given the fine-grained reasoning demands of this stage, we run inference with extended thinking enabled on Gemma-4-31B-it; the prompt is provided in Figures~\ref{fig:prompt_relevant_chunk-1} and \ref{fig:prompt_relevant_chunk-2}. 

\noindent\textbf{(v) Annotation Validation.} We apply four joint acceptance criteria to remove annotations that passed earlier stages but fail closer scrutiny. The evidence must be sufficient, in that the resolving sentence supplies the exact missing information defined by the gap type. It must be non-inferential, in that no additional guessing or assumption is required to close the gap. It must support independence, in that the target utterance becomes unambiguously self-contained once the resolving information is incorporated. It must be non-redundant, in that the resolving content is meaningfully distinct from what is already present in $\mathcal{W}_t$. This step is specifically designed to catch two recurring failure modes, namely cases in which resolving evidence is in fact present in $\mathcal{W}_t$ but was overlooked in Stage~(ii), and cases in which the resolution is too vague or indirect to constitute a verifiable resolution. The prompt is provided in Figure~\ref{fig:prompt_relevant_chunk_validation}; we run all three LLMs, Qwen3.6-27B, Gemma-4-31B-it, and GLM-4.7-Flash, at temperature 0.0.

\section{Evaluation Metric Formulations}
\label{sec:appendix_metrics}


\noindent\textbf{Memory Recall.} 
Let $\mathcal{G}_t = \{g_1, \ldots, g_{|\mathcal{G}_t|}\}$ denote the set of annotated contextual gaps for step $t$, and let $\mathcal{G}^{+}_t \subseteq \mathcal{G}_t$ denote the subset for which $\mathcal{M}_t$ explicitly contains the resolving information, verified either verbatim or as a close semantic paraphrase of the supporting evidence. Memory recall is defined as 
\[
\mathrm{MemRecall}(\mathcal{M}_t, \mathcal{G}_t) = \frac{|\mathcal{G}^{+}_t|}{|\mathcal{G}_t|}. 
\]
This metric measures how completely the memory module preserves the historical evidence required for self-contained summarization, independently of the generated summary. All labels are produced by Qwen3.6-27B at temperature 0.0; the prompt is provided in Figure~\ref{fig:prompt_memory_recall}.


\smallskip \noindent\textbf{Window Completeness.} 
Let $\mathcal{U}^{\mathrm{sal}}_t = \{u_1, \ldots, u_{|\mathcal{U}^{\mathrm{sal}}_t|}\}$ denote the set of salient utterances in $\mathcal{W}_t$, and let $\mathcal{U}^{+}_t \subseteq \mathcal{U}^{\mathrm{sal}}_t$ denote the subset whose primary claim, comprising both the core referent and the essential relation, is recoverable from $S_t$. Window completeness is defined as 
\[
\mathrm{WinComp}(\mathcal{U}^{\mathrm{sal}}_t, S_t) = \frac{|\mathcal{U}^{+}_t|}{|\mathcal{U}^{\mathrm{sal}}_t|}.
\]
Narrative coverage of the same topic without recovering the specific referent and relation does not qualify. We use Qwen3.6-27B (temperature 0.0) for this stage; see Figure~\ref{fig:prompt_comp} for the prompt.


\smallskip \noindent\textbf{Gap-Resolution Completeness.} 
Let $\mathcal{G}^{++}_t \subseteq \mathcal{G}^{+}_t$ denote the subset of memory-resolved gaps whose minimal gap-resolving fact, including both the specific referent and the essential relation, is explicitly expressed in $S_t$. Gap-resolution completeness is defined as 
\[
\mathrm{GapComp}(\mathcal{G}^{+}_t, S_t) = \frac{|\mathcal{G}^{++}_t|}{|\mathcal{G}^{+}_t|}. 
\]
This metric measures how well the model utilizes recovered evidence in generation, conditioned on successful memory retrieval. The prompt is provided in Figures~\ref{fig:prompt_gcomp_1} and~\ref{fig:prompt_gcomp_2}; we run Qwen3.6-27B at temperature 0.0.


\smallskip \noindent\textbf{Conciseness.}
Let $\mathcal{S}^{\mathrm{win}}_t \subseteq S_t$ denote the set of summary sentences that cover at least one salient utterance, as determined by $\mathcal{U}^{+}_t$, and let $\mathcal{S}^{\mathrm{gap}}_t \subseteq S_t$ denote the set of summary sentences that explicitly recover at least one gap-resolving fact, as determined by $\mathcal{G}^{++}_t$. Conciseness is defined as 
\[
\mathrm{Conciseness}(S_t) = \frac{|\mathcal{S}^{\mathrm{win}}_t \cup \mathcal{S}^{\mathrm{gap}}_t|}{|S_t|}. 
\]
This metric measures the proportion of summary sentences that serve a demonstrable communicative function, penalizing content that neither reflects salient window utterances nor expresses resolved contextual gaps. Conciseness shares the evaluation outputs of window completeness and gap-resolution completeness and requires no additional model call.

\smallskip
\noindent\textbf{Faithfulness.} 
Let $S_t = \{s_1, \ldots, s_N\}$ be the generated summary of $N$ sentences. 
Unlike completeness and conciseness, which are measured at the sentence level, faithfulness is evaluated at the atomic fact level. Summaries of $\mathcal{W}_t$ that incorporate resolved contextual gaps tend to produce long sentences that conflate multiple verifiable claims; sentence-level verification is therefore too coarse to detect partial hallucinations within a single sentence. 
Following \citet{min2023factscore} and \citet{wei2024long}, each sentence $s_n$ is first decomposed into a set of atomic facts $\mathcal{F}_n = \{f_{n,1}, \ldots, f_{n,|\mathcal{F}_n|}\}$, each expressing a single self-contained verifiable claim. 
Each atomic fact is then verified following the fact-checking protocol of \citet{song2024finesure}, extended to support multi-section dialogue transcripts. $\mathcal{F}^{*}_n \subseteq \mathcal{F}_n$ denotes the subset verified as factually grounded, and faithfulness is defined as
\[
\mathrm{Faithfulness}(S_t) = \frac{\sum_{n=1}^{N} |\mathcal{F}^{*}_n|}{\sum_{n=1}^{N} |\mathcal{F}_n|}. 
\] 
The verification source differs by method type. For memory-based methods, each atomic fact is verified against $\mathcal{M}_t \cup \mathcal{W}_t$. For full-context methods, $\mathcal{H}_t$ is too long to serve directly as a grounding document; we therefore apply sparse retrieval~\citep{robertson2009probabilistic} over $\mathcal{H}_t$ to select the top-50 chunks most relevant to each atomic fact, and verify against this retrieved set using the same prompt. Both steps are implemented with Qwen3.6-27B at temperature 0.0; the decomposition and verification prompts are provided in Figures~\ref{fig:prompt_atomic} and~\ref{fig:prompt_fact}.

\newcommand{\oc}[1]{\cellcolor{oraclerow}#1}
\newcommand{\pc}[1]{\cellcolor{proposedrow}#1}

\begin{table*}[t!]
\scriptsize
    \centering

\setlength{\tabcolsep}{1.8pt}
\begin{tabular}{X{0.9cm}L{2.0cm}X{0.82cm}X{0.82cm}X{0.82cm}X{0.82cm}X{1.4cm}X{0.82cm}X{0.82cm}X{0.82cm}X{0.82cm}X{1.4cm}}

\toprule
\multirow[c]{2}{*}{\makecell{Summ.\\Model}} & \multirow[c]{2}{*}{Memory Type} & \multicolumn{5}{c}{Chit-Chat} & \multicolumn{5}{c}{Workplace} \\
\cmidrule(lr){3-7} \cmidrule(l){8-12}
 &  & Win-Comp  & Gap-Comp & Conc & Faith & Composite & Win-Comp  & Gap-Comp & Conc & Faith & Composite \\
 \midrule
\multirow{10}{*}{\rotatebox{90}{\parbox{2.6cm}{\centering Qwen3.5-4B}}} & \oc{Ideal Memory} & \oc{0.7185} & \oc{0.8141} & \oc{0.7032} & \oc{0.7857} & \oc{0.7554} & \oc{0.6514} & \oc{0.8375} & \oc{0.7576} & \oc{0.8350} & \oc{0.7712} \\
& \oc{No Memory} & \oc{0.7186} & \oc{N/A} & \oc{0.6433} & \oc{0.8707} & \oc{0.5582} & \oc{0.6517} &  \oc{N/A} & \oc{0.6900} & \oc{0.8819} & \oc{0.5559} \\
& \oc{Full Memory} & \oc{0.3135} & \oc{0.4566} & \oc{0.2691} & \oc{0.6607} & \oc{0.4250} & \oc{0.1811} & \oc{0.3378} & \oc{0.2124} & \oc{0.8076} & \oc{0.3843} \\
 & Recent Memory & 0.6280 & 0.1989 & 0.5076 & 0.7895 & 0.4134 & 0.6334 & 0.3526 & 0.6093 & 0.8460 & 0.6103 \\
 & Inc. Summary & 0.6551 & 0.2681 & 0.5084 & 0.6568 & 0.5221 & 0.6789 & 0.2093 & 0.6391 & 0.6721 & 0.5499 \\
 & Hier. Summary & 0.6989 & 0.3321 & 0.5307 & 0.6381 & 0.5499 & 0.6599 & 0.1367 & 0.6168 & 0.6124 & 0.5064 \\
 & Sparse& 0.6482 & 0.2500 & 0.5374 & 0.7734 & 0.5518 & 0.6464 & 0.4201 & 0.6606 & 0.8189 & 0.6363 \\
 & Dense & 0.6527 & 0.3033 & 0.5471 & 0.7790 & 0.5705 & 0.6182 & 0.3634 & 0.6604 & 0.8002 & 0.6106 \\
 & Hybrid & 0.6404 & 0.2884 & 0.5494 & 0.8099 & 0.5720 & 0.6331 & 0.4163 & 0.6629 & 0.8569 & 0.6423 \\
 \cmidrule{2-12}
  & \pc{\algname{}} & \pc{0.6913} & \pc{0.4559} & \pc{0.6038} & \pc{0.7853} & \pc{0.6341} & \pc{0.6684} & \pc{0.5165} & \pc{0.6900} & \pc{0.8255} & \pc{0.6751} \\
\midrule
\multirow{10}{*}{\rotatebox{90}{\parbox{2.6cm}{\centering Qwen3.5-9B}}} & \oc{Ideal Memory} & \oc{0.7521} & \oc{0.7773} & \oc{0.7322} & \oc{0.8685} & \oc{0.7825} & \oc{0.7004} & \oc{0.7877} & \oc{0.8066} & \oc{0.8830} & \oc{0.7946} \\
 & \oc{No Memory} & \oc{0.7033} & \oc{N/A} & \oc{0.6889} & \oc{0.9043} & \oc{0.5741} & \oc{0.6408} & \oc{N/A} & \oc{0.7306} & \oc{0.8963} & \oc{0.5669} \\
 & \oc{Full Memory} & \oc{0.4421} & \oc{0.4429} & \oc{0.4063} & \oc{0.7071} & \oc{0.4996} & \oc{0.2176} & \oc{0.3259} & \oc{0.2968} & \oc{0.7674} & \oc{0.4019} \\
 & Recent Memory & 0.7147 & 0.1937 & 0.6013 & 0.8668 & 0.5941 & 0.6784 & 0.3382 & 0.7252 & 0.8831 & 0.6562 \\
 & Inc. Summary & 0.7451 & 0.3076 & 0.6436 & 0.7072 & 0.6009 & 0.7085 & 0.1939 & 0.7378 & 0.6969 & 0.5843 \\
 & Hier. Summary & 0.7810 & 0.3092 & 0.6461 & 0.6968 & 0.6083 & 0.7016 & 0.1229 & 0.7125 & 0.6674 & 0.5511 \\
 & Sparse & 0.7373 & 0.2695 & 0.6358 & 0.8621 & 0.6262 & 0.6892 & 0.4086 & 0.7571 & 0.8774 & 0.6831 \\
 & Dense & 0.7508 & 0.3401 & 0.6678 & 0.8812 & 0.6600 & 0.6777 & 0.3583 & 0.7399 & 0.8750 & 0.6627 \\
 & Hybrid & 0.7439 & 0.3192 & 0.6277 & 0.8600 & 0.6377 & 0.6944 & 0.3986 & 0.7614 & 0.8890 & 0.6859 \\
 \cmidrule{2-12}
 & \pc{\algname{}} & \pc{0.7501} & \pc{0.4122} & \pc{0.6757} & \pc{0.8624} & \pc{0.6751} & \pc{0.7030} & \pc{0.4762} & \pc{0.7717} & \pc{0.8857} & \pc{0.7091} \\
\midrule
\multirow{10}{*}{\rotatebox{90}{\parbox{2.6cm}{\centering Gemma-4-E2B-it}}}  & \oc{Ideal Memory} & \oc{0.7845} & \oc{0.6078} & \oc{0.4653} & \oc{0.9419} & \oc{0.6999} & \oc{0.7381} & \oc{0.6654} & \oc{0.5891} & \oc{0.9682} & \oc{0.7402} \\
& \oc{No Memory} & \oc{0.7265} & \oc{N/A} & \oc{0.4813} & \oc{0.9521} & \oc{0.5400} & \oc{0.6413} & \oc{N/A} & \oc{0.5752} & \oc{0.9741} & \oc{0.5476} \\
& \oc{Full Memory} & \oc{0.5029} & \oc{0.4002} & \oc{0.2322} & \oc{0.8215} & \oc{0.4892} & \oc{0.1178} & \oc{0.2918} & \oc{0.0872} & \oc{0.8492} & \oc{0.3365} \\
 & Recent Memory & 0.7179 & 0.1609 & 0.3615 & 0.9387 & 0.5445 & 0.6858 & 0.3278 & 0.5235 & 0.9673 & 0.6261 \\
 & Inc. Summary & 0.8031 & 0.2236 & 0.3870 & 0.8238 & 0.5593 & 0.7704 & 0.1569 & 0.5400 & 0.8810 & 0.4636 \\
 & Hier. Summary & 0.8103 & 0.2286 & 0.3908 & 0.8313 & 0.5194 & 0.7584 & 0.1012 & 0.5266 & 0.8847 & 0.5871 \\
 & Sparse & 0.7478 & 0.1998 & 0.3846 & 0.9419 & 0.5689 & 0.7094 & 0.3676 & 0.5536 & 0.9707 & 0.6503 \\
 & Dense & 0.7555 & 0.2806 & 0.3992 & 0.9415 & 0.5942 & 0.6887 & 0.3360 & 0.5506 & 0.9657 & 0.6353 \\
 & Hybrid & 0.7566 & 0.2394 & 0.4076 & 0.9381 & 0.5854 & 0.7079 & 0.3986 & 0.5512 & 0.9689 & 0.6566 \\
 \cmidrule{2-12}
 & \pc{ \algname{} } & \pc{0.7854 } & \pc{0.3133 } & \pc{0.4089 } & \pc{0.9401 } & \pc{0.6119 } & \pc{0.7529 } & \pc{0.4230 } & \pc{0.5632 } & \pc{0.9681 } & \pc{0.6768} \\
\bottomrule
\end{tabular}
\caption{Summary quality by domain and summarization model across memory construction methods. Win-Comp, Gap-Comp, Conc, Faith, and Composite denote window completeness, gap-resolution completeness, conciseness, faithfulness and their mean, respectively.}
\label{tab:summary-model-domain}
\end{table*}

\begin{table}[t!]
\scriptsize
    \centering

\setlength{\tabcolsep}{5pt}
\begin{tabular}{L{2cm}X{2cm}X{2cm}}

\toprule
Memory Type & Chit-Chat & Workplace \\
\midrule
Recent Memory & 0.3019 & 0.4679  \\
Inc. Summary & 0.4859 & 0.2942 \\
Hier. Summary & 0.5519 & 0.2059  \\
Sparse & 0.4057 & 0.5666 \\
Dense & 0.4983 & 0.4854 \\
Hybrid & 0.4798 & 0.5719  \\
\midrule
\rowcolor{proposedrow} \algname{}& 0.6918 & 0.7018 \\
\bottomrule
\end{tabular}
\caption{Domain-wise Memory Recall scores across memory construction methods, reported separately for Chit-Chat and Workplace domains.}
\label{tab:domain-wise-avg-quality}
\end{table}

\begin{table}[t!]
\scriptsize
    \centering

\setlength{\tabcolsep}{3pt}
\begin{tabular}{L{3.0cm}X{2.0cm}X{1.6cm}X{2.0cm}X{1.6cm}}
\toprule
\multirow[c]{2}{*}{Model} & \multicolumn{2}{c}{Chit-Chat} & \multicolumn{2}{c}{Workplace} \\
\cmidrule(lr){2-3} \cmidrule(l){4-5}
 & Memory Recall & Gap-Comp & Memory Recall & Gap-Comp \\
\midrule
Qwen3.5-9B & 0.6810 & 0.4006 & 0.6322 & 0.4581 \\
Gemma-4-E2B-it & 0.6509 & 0.4041 & 0.6361 & 0.4652 \\
Qwen3.5-4B & 0.6918 & 0.3938 & 0.7018 & 0.4719\\
\bottomrule
\end{tabular}
\caption{Gap detection model variation. Domain-wise memory recall and Gap-Comp scores are reported. Summaries are generated with Qwen3.5-4B across all configurations. Gap-Comp denotes Gap-Resolution Completeness.}
\label{tab:gap_detection_model_ablation}
\end{table}
\section{Additional Analysis of \algname{}}
\label{sec:appendix_remember}
This appendix provides additional analyses of \algname{} beyond the main results. We examine whether the gains are consistent across different summarization models and dialogue domains, whether memory recall improvements hold separately in chit-chat and workplace settings, and whether gap detection depends on the choice of compact LLM. Together, these analyses test whether the benefits of gap-conditioned memory construction are robust to generator choice, domain characteristics, and the model used for detecting unresolved contextual gaps.

\subsection{Gap Detection Model Ablation}
\label{sec:appendix_gep_detection_model_ablation}
Table~\ref{tab:gap_detection_model_ablation} reports the effect of varying the gap detection model across two domains. All three 
compact models yield comparable performance, indicating that gap detection does not require a large model. Performance variation across domains is more pronounced than across models, suggesting that domain characteristics drive recall differences more than model capacity.
  
\subsection{Memory Recall Across Domains}
\label{sec:appendix_mr_domain_wise}
Table~\ref{tab:domain-wise-avg-quality} reports Memory Recall separately for the chit-chat and workplace domains. Across both domains, \algname{} achieves the highest recall among memory construction baselines, reaching 0.6918 in chit-chat and 0.7018 in workplace. This indicates that gap-conditioned memory construction improves evidence recovery across both dialogue settings.
The domain-wise results show different baseline behavior. In chit-chat, summarization-based memory is relatively strong, with \texttt{Hier.\,Summary} reaching 0.5519 recall and \texttt{Inc.\,Summary} reaching 0.4859. This is likely because chit-chat dialogues are shorter on average, making compression less lossy than in longer workplace histories. In workplace dialogues, summarization-based memory degrades sharply, while retrieval-based memory is stronger, with \texttt{Hybrid} reaching 0.5719 recall. This suggests that as dialogue histories become longer, repeated compression loses fine-grained evidence, whereas retrieval better preserves localized historical evidence. 
Despite these domain differences, \algname{} consistently improves over the strongest baseline in each domain. It improves over \texttt{Hier.\,Summary} by 0.1399 in chit-chat and over \texttt{Hybrid} by 0.1299 in workplace. These results show that explicitly querying unresolved contextual gaps recovers evidence that both compression-based and window-level retrieval baselines miss.

\begin{table*}[t!]
\scriptsize
\centering
\setlength{\tabcolsep}{2.5pt}
\renewcommand{\arraystretch}{1.25}
\begin{tabular}{
@{}
L{1.1cm}
L{3.1cm}
L{4.6cm}
C{0.95cm}
C{0.9cm}
C{1.2cm}
C{0.85cm}
@{}
}
\toprule
Type & Name / Usage & Source / ID & Params & VRAM & \makecell{GPU h\\/ call} & \makecell{Sec.\\/ call} \\
\midrule
\multirow{7}{*}{LLM}
& Qwen3.5-4B & \idlink{https://huggingface.co/Qwen/Qwen3.5-4B}{Qwen/\brk Qwen3.5-4B} & 4B & 30GB & 0.000436 & 1.57 \\
& Qwen3.5-9B & \idlink{https://huggingface.co/Qwen/Qwen3.5-9B}{Qwen/\brk Qwen3.5-9B} & 9B & 60GB & 0.000392 & 1.41 \\
& Qwen3.6-27B & \idlink{https://huggingface.co/Qwen/Qwen3.6-27B}{Qwen/\brk Qwen3.6-27B} & 27B & 90GB & 0.001060 & 3.81 \\
& Gemma-4-E2B-it & \idlink{https://huggingface.co/google/gemma-4-E2B-it}{google/\brk gemma-4-\brk E2B-it} & 5B & 30GB & 0.000387 & 1.39 \\
& Gemma-4-31B-it & \idlink{https://huggingface.co/google/gemma-4-31B-it}{google/\brk gemma-4-\brk 31B-it} & 31B & 90GB & 0.000458 & 1.65 \\
& \hspace{1.2mm}\textit{+ thinking mode} & & 31B & 90GB & 0.020000 & 72.00 \\
& GLM-4.7-Flash & \idlink{https://huggingface.co/zai-org/GLM-4.7-Flash}{zai-org/\brk GLM-4.7-\brk Flash} & 30B & 90GB & 0.000162 & 0.58 \\
\midrule
Embed.
& Qwen3-Embedding-0.6B & \idlink{https://huggingface.co/Qwen/Qwen3-Embedding-0.6B}{Qwen/\brk Qwen3-\brk Embedding-0.6B} & 0.6B & 3GB & -- & -- \\
\midrule
\multirow{2}{*}{Preproc.}
& Sentence segmentation & \texttt{NLTK punkt}~\cite{bird2009nltk} & -- & -- & -- & -- \\
& Token counting & \idlink{https://github.com/openai/tiktoken}{tiktoken cl100k\_\brk base} & -- & -- & -- & -- \\
\midrule
\multirow{3}{*}{Datasets}
& EverMemBench & \idlink{https://huggingface.co/datasets/EverMind-AI/EverMemBench-Dynamic}{EverMind-AI/\brk EverMemBench-\brk Dynamic} & -- & -- & -- & -- \\
& LoCoMo & \idlink{https://github.com/snap-research/locomo}{snap-research/\brk locomo} & -- & -- & -- & -- \\
& RealTalk21 & \idlink{https://github.com/danny911kr/REALTALK}{danny911kr/\brk REALTALK} & -- & -- & -- & -- \\
\bottomrule
\end{tabular}
\caption{Model, embedding, preprocessing, and dataset implementation details. All model runs used a single NVIDIA H200 GPU at FP16 precision. VRAM denotes the approximate peak allocated GPU memory. GPU hours and seconds report the average cost per model call, computed over all calls made with each model.}
\label{tab:model_implementation}
\end{table*}

\subsection{Performance Across Summarizers and Domains}
\label{sec:appendix_summarizer_wise}

Table~\ref{tab:summary-model-domain} reports summary quality across three summarization models and two dialogue domains. Across all summarizers, \algname{} outperforms the memory construction baselines in both domains, achieving the highest composite score in every setting, indicating the best overall balance among window completeness, gap-resolution completeness, conciseness, and faithfulness. The gains are mainly driven by gap-resolution completeness. \algname{} consistently improves Gap-Comp over recency-, summarization-, and retrieval-based memory, showing that the recovered evidence is not only present in memory but also usable by different downstream summarizers.

The results also show that the benefit of \algname{} is not tied to a particular generator. With Qwen3.5-4B~\citep{qwen3.5}, Qwen3.5-9B, and Gemma-4-E2B-it~\citep{gemma4}, \algname{} yields the strongest overall performance across both domains while maintaining competitive window completeness and faithfulness. Retrieval-based methods~(\texttt{Sparse}, \texttt{Dense}, \texttt{Hybrid}) often preserve faithfulness, but their Gap-Comp remains lower because window-level retrieval misses evidence required to resolve underspecified current-window utterances. Summarization-based memory~(\texttt{Inc.\,Summary}, \texttt{Hier.\,Summary}) sometimes improves window completeness, but its Gap-Comp remains limited, indicating that repeated compression discards fine-grained historical evidence.

The diagnostic references further clarify the bottleneck. \texttt{No Memory} preserves high faithfulness but cannot resolve history-dependent gaps, while \texttt{Full Memory} often lowers window completeness and conciseness, especially in the workplace domain. \texttt{Ideal Memory} provides a reference-evidence upper bound. Its gap-resolution scores remain below perfect, indicating that even when relevant evidence is supplied, summary generation can still fail to fully express the resolved information. Overall, these results suggest that gap-conditioned memory construction improves overall summary quality consistently across summarization models and dialogue domains.

\section{Model and Hardware Details}
\label{sec:appendix_model_implementation}
Table~\ref{tab:model_implementation} summarizes the implementation resources used in our experiments, including model and embedding identifiers, dataset sources, hardware, precision, peak allocated VRAM, and measured average runtime cost. 
We use Qwen3.5-4B~\citep{qwen3.5}, Qwen3.5-9B~\citep{qwen3.5}, and Gemma-4-E2B-it~\citep{gemma4} for gap detection and summary generation. We use Qwen3.6-27B~\citep{qwen3.6-27b}, Gemma-4-31B-it~\citep{gemma4}, and GLM-4.7-Flash~\citep{5team2025glm45agenticreasoningcoding} for benchmark construction and evaluation. 
For Gemma-4-31B-it, we report non-thinking mode by default, with thinking-mode costs shown in parentheses.
We use NLTK~\citep{bird2009nltk} \texttt{punkt} for sentence segmentation and \texttt{tiktoken}~\citep{tiktoken} with the \texttt{cl100k\_base} encoding for approximate token counting when constructing windows, chunks, and memory budgets. All model runs use a single NVIDIA H200 GPU. All results are reported from a single experimental run.

\section{Scientific Artifacts}
\label{sec:appendix_scientific_artifacts}
All experiments use existing research datasets and open-weight language models. The model checkpoints are publicly accessible through Hugging Face. Benchmark statistics and implementation details are provided in Appendix~\ref{sec:appendix_dataset} and Appendix~\ref{sec:appendix_model_implementation}.

\section{Use of Generative AI}
We used AI assistants for coding assistance~(Codex) and for grammar checking~(Claude) during the preparation of this paper.

\begin{figure*}[t]
\centering
\begin{promptfigbox}{Prompt: Gap Detection in \algname{}}

You are building retrieval queries for long-dialogue memory.

\vspace{2pt}
\noindent Input is ONLY the current conversation window. Earlier history is hidden.

\vspace{2pt}
\noindent Task:

Find current-window utterances that a concise summary would mention, but that
cannot be stated faithfully without earlier dialogue.

Do not answer the missing context. Convert each missing slot into a retrieval
need for earlier dialogue.

\vspace{2pt}
\noindent Select an utterance when:
\begin{promptitems}
\item it is a concrete action, plan, decision, result, request, question, issue,
artifact, task, review, test, fix, status update, owner assignment, or
personal update;
\item it would likely appear in a summary of the current window;
\item a specific referent, artifact, owner, project, scope, version, rationale,
requirement, decision, result, deadline, or prior state is missing from the
current window;
\item the missing context is likely stated in earlier dialogue.
\end{promptitems}

\noindent Do not select greetings, thanks, small talk, generic reactions, self-contained
utterances, image-only references, or merely nice-to-know details.

\vspace{2pt}
\noindent For each gap:
\begin{promptitems}
\item quote the exact current-window utterance or the shortest complete span;
\item gap\_reason must name the missing slot and why the quote is not self-contained;
\item evidence\_need\_query starts with "retrieve earlier dialogue ..." and uses only
visible words from the current window plus the missing-slot type;
\item keywords are stable anchor terms copied from the quote or surrounding topic;
\item never guess the answer.
\end{promptitems}

\noindent Prefer recall for real summary-worthy gaps. If several related utterances refer
to the same missing context, select the most informative one.

\vspace{2pt}
\noindent Output up to \texttt{\{max\_gaps\}} gaps. Return compact JSON only:

\vspace{1pt}
{\ttfamily\scriptsize
\noindent\{
\newline
\hspace*{1em}"gaps": [
\newline
\hspace*{2em}\{
\newline
\hspace*{3em}"utterance\_index": 0,
\newline
\hspace*{3em}"quote": "exact quote, \textless= 220 chars",
\newline
\hspace*{3em}"gap\_reason": "missing slot and why earlier context is needed, \textless= 220 chars",
\newline
\hspace*{3em}"evidence\_need\_query": "retrieve earlier dialogue identifying/stating/explaining ...",
\newline
\hspace*{3em}"keywords": ["up", "to", "8", "anchors"]
\newline
\hspace*{2em}\}
\newline
\hspace*{1em}]
\newline
\}
}

\vspace{2pt}
\noindent\textbf{Positive examples:}

\vspace{1pt}
\noindent Utterance: "Please help review it again to see if the logic is sound."

\noindent Output gap\_reason: "The artifact referred to by 'it' and the logic being reviewed are not identified in the current window."

\noindent Output evidence\_need\_query: "retrieve earlier dialogue identifying the artifact to review and the logic that should be checked"

\vspace{2pt}
\noindent Utterance: "Received, I will immediately arrange test cases, and formal testing will begin tomorrow."

\noindent Output gap\_reason: "The feature or task being tested is not identified in the current window."

\noindent Output evidence\_need\_query: "retrieve earlier dialogue identifying the feature or task for the test cases and formal testing"

\vspace{2pt}
\noindent Utterance: "This confirms we should not change the current allocation."

\noindent Output gap\_reason: "The allocation and confirming rationale are not stated in the current window."

\noindent Output evidence\_need\_query: "retrieve earlier dialogue stating the current allocation and why it should remain unchanged"

\vspace{2pt}
\noindent\textbf{Negative examples:}

\vspace{1pt}
\noindent Utterance: "Thanks, that sounds great!"

\noindent Do not output: not summary-worthy and no required earlier evidence.

\vspace{2pt}
\noindent Utterance: "The API documentation was updated today."

\noindent Do not output if the current window already identifies which API or the summary
can faithfully state the update without earlier context.

\vspace{2pt}
\noindent\textbf{Current window:}

\noindent\texttt{\{window\}}

\end{promptfigbox}

\caption{Prompt template for gap detection in \algname{}.}
\label{fig:prompt_remember}
\end{figure*}
\begin{figure*}[t]
\centering
\begin{promptfigbox}{Prompt: Memory-Augmented Dialogue Summary Generation}

Your task is to generate a 'Reinforced Dialogue Summary' by synthesizing current [Window] with [Memory] in a fact-centered manner.

\vspace{2pt}
\noindent\textbf{Instructions:}

\begin{promptenum}
\item Fully understand [Window] then carefully read [Memory].
\item Map all entities in [Window] to their specific Identity, Role, or Context defined in [Memory]. Replace all vague terms with these precise identifiers.
\item Enrich the content of [Window] by integrating missing context such as coreference, causality, logical continuity, and technical details from [Memory] where applicable.
\item Generate a concise, high-density, and contextually reinforced summary \textbf{centered on the content in [Window]}. The result should be a cohesive narrative that reflects the core insights and the logical evolution of the situation.
\end{promptenum}

\noindent Provide your output as JSON. No additional explanation.

\vspace{1pt}
\noindent{\ttfamily\scriptsize
\{\{"summary": "Your high-density, contextually reinforced summary here."\}\}
}

\vspace{2pt}
\noindent\textbf{[Window]}

\noindent\texttt{\{window\}}

\vspace{2pt}
\noindent\textbf{[Memory]}

\noindent\texttt{\{memory\}}

\end{promptfigbox}
\caption{Prompt template for memory-augmented dialogue summary generation.}
\label{fig:prompt_summary_generation_w_memory}
\end{figure*}
\begin{figure*}[t]
\centering
\begin{promptfigbox}{Prompt: Memory Recall Evaluation}

You are a strict memory coverage evaluator.

Your task is to judge whether MEMORY explicitly contains the information
needed to resolve each GAP.

For this task, the information needed to resolve a GAP is already defined in
the GAP ITEM. You must not redefine it.

\begin{promptitems}
\item supporting\_sentences provide the information that resolves the GAP.
\item supporting\_reasoning explains why that information resolves the GAP.
\end{promptitems}

\noindent Do NOT decide that the GAP needs additional information beyond the provided
supporting\_sentences and supporting\_reasoning. Your job is not to redesign the
gap resolution. Your job is only to check whether MEMORY contains that key
information explicitly, either verbatim or as a close semantic paraphrase.

\vspace{2pt}
\noindent For each GAP:

\begin{promptenum}
\item Identify the key information expressed by supporting\_sentences, using
supporting\_reasoning only to understand why that information resolves the GAP.
\item Check whether that same key information is present in MEMORY.
\item Decide 1 if MEMORY contains that information explicitly, either verbatim or
as a close semantic paraphrase.
\item Decide 0 only if MEMORY does not contain the key information from the
supporting\_sentences, contains only broad topical overlap, or contradicts it.
\end{promptenum}

\vspace{2pt}
\noindent Important:

\begin{promptitems}
\item The coverage criterion is fixed by supporting\_sentences and
supporting\_reasoning, not by your own idea of what else might be useful to
resolve the salient\_utterance.
\item Do NOT introduce new required information.
\item Do NOT mark 0 because some additional background, definition, detail, or
rationale would be helpful if it is not part of supporting\_sentences or
supporting\_reasoning.
\item Use DIALOGUE only to understand the salient\_utterance context. Do NOT derive
additional required information from DIALOGUE.
\item The final decision must be based only on MEMORY, but the target information to
look for is defined by supporting\_sentences and supporting\_reasoning.
\item MEMORY does not need to quote supporting\_sentences verbatim. A close semantic
paraphrase is enough, but broad topical similarity is not enough.
\item Do not use outside knowledge or common sense.
\item If MEMORY contains the key information from supporting\_sentences, choose 1.
\end{promptitems}

\vspace{2pt}
\noindent Gap-type guide:

\begin{promptitems}
\item Coreference (C): MEMORY must contain the information that identifies
the referent.
\item Attribute (A): MEMORY must contain the information for the missing
attribute, state, role, scope, constraint, or specification.
\item Logic (L): MEMORY must contain the information for the rationale,
causal link, motivation, dependency, or justification.
\end{promptitems}

\vspace{2pt}
\noindent Return JSONL only:

\vspace{1pt}
{\ttfamily\scriptsize
\noindent\{\{
\newline
\hspace*{1em}"gap\_id": "...",
\newline
\hspace*{1em}"gap\_type": "C/A/L",
\newline
\hspace*{1em}"target\_information": [
\newline
\hspace*{2em}"Atomic key information from supporting\_sentences, interpreted with supporting\_reasoning"
\newline
\hspace*{1em}],
\newline
\hspace*{1em}"evidence\_from\_memory": [
\newline
\hspace*{2em}\{\{
\newline
\hspace*{3em}"target\_item": "...",
\newline
\hspace*{3em}"memory\_index": "M\# or null",
\newline
\hspace*{3em}"text": "Exact or semantically equivalent supporting span from MEMORY"
\newline
\hspace*{2em}\}\}
\newline
\hspace*{1em}],
\newline
\hspace*{1em}"missing\_target\_information": [
\newline
\hspace*{2em}"Key target information not found in MEMORY"
\newline
\hspace*{1em}],
\newline
\hspace*{1em}"decision": "1 or 0",
\newline
\hspace*{1em}"reasoning": "One concise sentence explaining whether MEMORY contains the target information."
\newline
\}\}
}

\vspace{2pt}
\noindent\textbf{[DIALOGUE]}

\noindent\texttt{\{window\}}

\vspace{2pt}
\noindent\textbf{[MEMORY]}

\noindent\texttt{\{memory\}}

\vspace{2pt}
\noindent\textbf{[GAP ITEMS]}

\noindent\texttt{\{gaps\}}

\end{promptfigbox}

\caption{Prompt template for memory recall evaluation.}
\label{fig:prompt_memory_recall}
\end{figure*}
\begin{figure*}[t]
\centering
\begin{promptfigbox}{Prompt: Window Completeness Evaluation}

For each salient utterance (U\#), follow these steps:

\begin{promptenum}
\item Identify the single most important claim in the utterance. Decompose it into two parts:

\noindent (a) the core referent --- the specific entity, event, decision, or quantity at the center of the claim;

\noindent (b) the essential relation --- the role this referent plays (who did what, when, why, or to whom) that makes the utterance salient.

\item Decide whether a reader who sees only the summary could recover the claim with both (a) the correct referent and (b) the correct relation.

\begin{promptitems}
\item Paraphrase is acceptable for the referent (synonyms, role-based references, alternate phrasings) as long as it uniquely identifies the same thing in (a).
\item Narrative coverage of the same topic, project, or domain --- without naming the specific referent or stating the specific relation --- does NOT count. Score 0.0 in that case.
\item A summary that mentions the referent but only in a different context (different action, different time, different participants) does NOT count for that relation.
\item Do not use outside knowledge or commonsense inference. The claim must be directly recoverable from the summary text.
\end{promptitems}
\end{promptenum}

\vspace{2pt}
\noindent Scores (binary):

\begin{promptitems}
\item 1.0: The summary states both (a) the specific referent and (b) the specific relation. A reader recovers the complete claim from the summary alone.
\item 0.0: Anything else --- the referent is missing or only described by a generic role; the relation is generalized, vague, or missing; the summary only mentions the broader topic/narrative; the summary contradicts the claim; or the claim is only implied through commonsense. Partial coverage does NOT earn credit.
\end{promptitems}

\vspace{2pt}
\noindent Output JSON only. No explanation.

\vspace{1pt}
{\ttfamily\scriptsize
\noindent[
\newline
\hspace*{1em}\{\{
\newline
\hspace*{2em}"utterance\_num": "U0",
\newline
\hspace*{2em}"score": 1.0,
\newline
\hspace*{2em}"line\_number": ["S1", "S3"]
\newline
\hspace*{1em}\}\},
\newline
\hspace*{1em}...
\newline
]
}

\vspace{2pt}
\noindent\textbf{\{N\} Summary Sentences (S\#):}

\noindent\texttt{\{summary\}}

\vspace{2pt}
\noindent\textbf{\{M\} Salient Utterances (U\#):}

\noindent\texttt{\{salient\_utterances\}}

\end{promptfigbox}

\caption{Prompt template for window completeness evaluation.}
\label{fig:prompt_comp}
\end{figure*}
\begin{figure*}[t]
\centering
\begin{promptfigbox}{Prompt: Gap-Resolution Completeness Evaluation, Part I}

For each gap case (G\#), evaluate whether the summary recovers the missing context.

\vspace{2pt}
\noindent Goal:

Reward the summary only if it contains the specific missing context from relevant\_chunk that is absent from salient\_utterance.
Do NOT reward the summary for merely covering the same event, topic, relationship, preference, plan, project, document, task, or broad narrative.

\vspace{2pt}
\noindent For each gap case (G\#), follow these steps:

\begin{promptenum}
\item Identify the exact gap in the salient\_utterance.

Write one concrete question Q that asks for the missing antecedent, referent, source, target, object, decision, time, quantity, or relation.
Q must be answerable only by using relevant\_chunk, not by using salient\_utterance alone.

\item Select exactly one gap-resolving fact F from relevant\_chunk.

F must be the minimal fact that directly answers Q.
F must satisfy ALL of the following:

\begin{promptitems}
\item (a) Directness: F resolves the exact missing information in salient\_utterance. Do not choose a related fact just because it appears in the summary.
\item (b) Core referent: F includes the specific entity, event, decision, document, person, date, quantity, or object that answers Q.
\item (c) Essential relation: F includes the role this referent plays (who did what, when, why, to whom, about what, or based on what).
\item (d) Novelty: F is not already lexically or semantically present in salient\_utterance. If the summary could be written using only salient\_utterance without reading relevant\_chunk, score 0.0.
\item (e) Minimality: F should not be a broad summary of relevant\_chunk. It should be the smallest fact needed to fill the gap.
\end{promptitems}

If relevant\_chunk contains multiple possible facts, choose the one that most directly fills the missing slot in salient\_utterance.
Do not choose an alternative fact merely because the summary mentions it.
Do not use "OR" facts. Pick one F.

\item Decide whether the summary alone contains F.

The reader sees only the summary, not relevant\_chunk and not salient\_utterance.
\end{promptenum}

Apply ALL tests. Any single failure -\textgreater{} score 0.0.

\vspace{2pt}
\noindent\textbf{T1. Exact gap coverage:}

Does the summary answer Q with the same F selected from relevant\_chunk?

\vspace{2pt}
\noindent\textbf{T2. Referent + relation:}

Does at least one cited summary sentence explicitly contain both the core referent and the essential relation?
If the summary mentions only the referent, or only a related action, score 0.0.

\vspace{2pt}
\noindent\textbf{T3. Fidelity:}

The summary's answer to Q must match relevant\_chunk. If it gives a different plausible answer, different participant, different object, different time, or different relation, score 0.0.

\vspace{2pt}
\noindent\textbf{T4. Anti-echo:}

If the summary only restates the salient\_utterance's topic, category, or already-known referent without adding F from relevant\_chunk, score 0.0.

\vspace{2pt}
\noindent\textbf{T5. Anti-narrative:}

If the summary only states that an event, conversation, relationship, preference, routine, trip, health issue, project, task, meeting, document, decision, or implementation happened, but does not recover the missing concrete context required by Q, score 0.0.

\vspace{2pt}
\noindent\textbf{T6. Specificity:}

Generic role mentions such as "they", "someone", "the friend", "the family member", "the place", "the thing", "the plan", "the issue", "the team", "the project", "the document", "the solution", or "the requirements" do not count unless they uniquely identify the same concrete referent and relation as F.

\vspace{2pt}
\noindent\textbf{T7. No outside knowledge:}

Do not use commonsense, world knowledge, or assumptions to bridge missing information. F must be directly recoverable from summary text.

\vspace{2pt}
\noindent Scores (binary):

\begin{promptitems}
\item 1.0: The summary explicitly recovers the selected F, including both the specific referent and essential relation, and the cited summary sentence(s) support it.
\item 0.0: Any test fails, no valid F exists, the summary is silent on Q, the summary only covers the broad narrative, the summary only echoes salient\_utterance, or the summary contradicts F.
\end{promptitems}

\end{promptfigbox}

\caption{Prompt template for gap-resolution completeness evaluation, Part I.}
\label{fig:prompt_gcomp_1}
\end{figure*}

\begin{figure*}[t]
\centering
\begin{promptfigbox}{Prompt: Gap-Resolution Completeness Evaluation, Part II}

Examples (illustrative, do NOT copy verbatim):

\vspace{2pt}
\noindent\textbf{Example A --- score 1.0 (personal plan)}

\noindent salient\_utterance: "I finally booked it for next Friday."

\noindent relevant\_chunk: "Maya said she wanted to celebrate her birthday at the small Thai place near the river because it has vegan options."

\noindent Q: What was booked for next Friday?

\noindent F: "Maya's birthday celebration at the small Thai place near the river"

\noindent summary contains: "the speaker booked Maya's birthday dinner at the riverside Thai restaurant for next Friday"

\noindent score: 1.0

\vspace{2pt}
\noindent\textbf{Example B --- score 0.0 (wrong answer)}

\noindent salient\_utterance: "He said that was the reason he stopped going."

\noindent relevant\_chunk: "Daniel stopped going to the climbing gym after he hurt his wrist during a bouldering class."

\noindent Q: Why did Daniel stop going?

\noindent F: "Daniel stopped going to the climbing gym because he hurt his wrist during a bouldering class"

\noindent summary contains: "Daniel stopped going because the membership became too expensive"

\noindent score: 0.0

\vspace{2pt}
\noindent\textbf{Example C --- score 0.0 (salient echo)}

\noindent salient\_utterance: "I made sure to avoid it when I cooked for her."

\noindent relevant\_chunk: "Nina is allergic to walnuts, but almonds and cashews are fine."

\noindent Q: What did the speaker avoid when cooking for Nina?

\noindent F: "walnuts, because Nina is allergic to them"

\noindent summary contains: "the speaker cooked carefully for Nina and avoided an ingredient"

\noindent score: 0.0

\vspace{2pt}
\noindent\textbf{Example D --- score 0.0 (broad narrative)}

\noindent salient\_utterance: "That place is why I want to go back in June."

\noindent relevant\_chunk: "When we visited Kyoto last spring, the quiet garden behind Nanzen-ji was Elena's favorite stop."

\noindent Q: Which place made the speaker want to go back in June?

\noindent F: "the quiet garden behind Nanzen-ji in Kyoto"

\noindent summary contains: "the conversation described a memorable previous trip and a desire to return"

\noindent score: 0.0

\vspace{2pt}
\noindent\textbf{Example E --- score 1.0 (work/domain case)}

\noindent salient\_utterance: "This task can be closed today."

\noindent relevant\_chunk: "Today is the last day for the platform's microservice architecture diagram review."

\noindent Q: Which task can be closed today?

\noindent F: "the platform's microservice architecture diagram review can be closed today"

\noindent summary contains: "the platform's microservice architecture diagram review was completed and closed today"

\noindent score: 1.0

\vspace{2pt}
\noindent Output JSON only. No explanation.

\vspace{1pt}
{\ttfamily\scriptsize
\noindent[
\newline
\hspace*{1em}\{\{
\newline
\hspace*{2em}"gap\_num": "G0",
\newline
\hspace*{2em}"utterance\_num": "U0",
\newline
\hspace*{2em}"question": "What concrete missing information is required?",
\newline
\hspace*{2em}"selected\_F": "The minimal fact from relevant\_chunk that directly answers the question.",
\newline
\hspace*{2em}"score": 1.0,
\newline
\hspace*{2em}"line\_number": ["S1", "S4"]
\newline
\hspace*{1em}\}\},
\newline
\hspace*{1em}...
\newline
]
}

\vspace{2pt}
\noindent\textbf{\{N\} Summary Sentences (S\#):}

\noindent\texttt{\{summary\}}

\vspace{2pt}
\noindent\textbf{\{M\} Gap Cases (G\#):}

\noindent\texttt{\{gap\_cases\}}

\end{promptfigbox}

\caption{Prompt template for gap-resolution completeness evaluation, Part II.}
\label{fig:prompt_gcomp_2}
\end{figure*}
\begin{figure*}[t]
\centering
\begin{promptfigbox}{Prompt: Summary Fact Decomposition for Faitfulness}

You are decomposing a generated summary into self-contained evaluable facts for factuality evaluation.

\vspace{2pt}
\noindent For each summary sentence (S\#), follow these steps:

\begin{promptenum}
\item Identify all verifiable claims in the sentence, not only the main claim.

A claim is verifiable if it can be checked against the dialogue context as true or false.

\item Decompose each claim into a self-contained evaluable fact. For each fact, preserve:

\noindent (a) the core referent --- the specific entity, event, decision, state, quantity, or issue at the center of the claim;

\noindent (b) the essential relation --- what happened to the referent, who did what, when, why, how, or to whom;

\noindent (c) support-relevant details --- details that affect verification, such as time, location, quantity, frequency, negation, comparison, condition, cause, or consequence.

\item Split coordinated or compressed claims when they express different verifiable information.

\begin{promptitems}
\item Split claims connected by "and", "but", "while", "because", "so", "after", "before", or similar relations when each part can be verified separately.
\item If a causal, temporal, contrastive, or decision relation is itself asserted, preserve it as a separate evaluable fact.
\item Do not drop details that change the truth condition of the claim.
\end{promptitems}

\item Make each fact self-contained.

\begin{promptitems}
\item Resolve pronouns and vague references using only the generated summary.
\item You may use surrounding summary sentences only to resolve references.
\item Do not use the source dialogue, retrieved memory, external knowledge, or commonsense inference.
\item Do not add information that is not explicitly stated in the generated summary.
\end{promptitems}

\item Exclude non-verifiable or subjective statements.

\begin{promptitems}
\item Do not extract vague judgments such as "the discussion was productive" unless the summary explicitly states verifiable evidence for them.
\item Do not create trivial facts such as "the team existed", "the issue existed", or "the box existed."
\item Each fact must be traceable to a specific source span in the generated summary.
\end{promptitems}
\end{promptenum}

\vspace{2pt}
\noindent Output JSON only. No explanation.

\noindent Return a compact JSON array. Each object must contain exactly these keys:

\begin{promptitems}
\item "sentence\_num": the source sentence id such as "S0"
\item "fact\_id": a stable id such as "S0-F0"
\item "fact": one self-contained evaluable fact
\end{promptitems}

\noindent Return [] if there are no verifiable facts.

\noindent After the closing ] stop immediately.

\vspace{1pt}
{\ttfamily\scriptsize
\noindent[
\newline
\hspace*{1em}\{\{
\newline
\hspace*{2em}"sentence\_num": "S0",
\newline
\hspace*{2em}"fact\_id": "S0-F0",
\newline
\hspace*{2em}"fact": "A self-contained evaluable fact."
\newline
\hspace*{1em}\}\}
\newline
]
}

\vspace{2pt}
\noindent\textbf{\{N\} Summary Sentences (S\#):}

\noindent\texttt{\{summary\}}

\end{promptfigbox}

\caption{Prompt template for decomposing summary sentences into atomic facts.}
\label{fig:prompt_atomic}
\end{figure*}
\begin{figure*}[t]
\centering
\begin{promptfigbox}{Prompt: Faitfulness Evaluation}

You will receive a transcript followed by a corresponding summary. Your task is to assess the factuality of each summary sentence across nine categories:

\begin{promptitems}
\item no error: the statement is factually consistent with the transcript. This includes paraphrase, generalization, aggregation, and reasonable inference that is entailed by or follows naturally from the transcript, even if it is not stated verbatim. Do not penalize a sentence merely because the wording differs or because it synthesizes information across multiple turns.
\item out-of-context error: the statement asserts a specific fact that is neither stated in nor reasonably inferable from the transcript (genuine fabrication). Do NOT use this category for legitimate paraphrase, summary-level abstraction, or inference that the transcript supports.
\item entity error: the primary arguments (or their attributes) of the predicate are wrong.
\item predicate error: the predicate in the summary statement is inconsistent with the transcript.
\item circumstantial error: the additional information (like location or time) specifying the circumstance around a predicate is wrong.
\item grammatical error: the grammar of the sentence is so wrong that it becomes meaningless.
\item coreference error: a pronoun or reference with wrong or non-existing antecedent.
\item linking error: error in how multiple statements are linked together in the discourse (for example temporal ordering or causal link).
\item other error: the statement contains any factuality error which is not defined here.
\end{promptitems}

\noindent The transcript may be split into several labeled sections (for example prior context / memory and the current dialogue window, or retrieved passages). Treat all sections together as a single body of ground-truth evidence: a summary sentence is supported as long as ANY section supports it. Never flag a sentence merely because its supporting evidence sits in the prior-context/memory portion rather than the current window.

\vspace{2pt}
\noindent Instruction:

First, compare each summary sentence with the transcript. Judge whether the claim is contradicted by or unsupportable from the transcript --- not whether it is restated word-for-word. Reasonable inference, paraphrase, and synthesis across turns are acceptable and should be labeled "no error". Only assign an error category when the sentence states something that conflicts with the transcript or asserts a specific fact the transcript does not support.

Second, provide a single sentence explaining which factuality error the sentence has.

Third, answer the classified error category for each sentence in the summary.

\vspace{2pt}
\noindent Provide your answer in JSON format. The answer should be a list of dictionaries whose keys are "sentence", "reason", and "category". No additional information:

\vspace{1pt}
{\ttfamily\scriptsize
\noindent[
\{\{"sentence": "first sentence", "reason": "your reason", "category": "no error"\}\},
\{\{"sentence": "second sentence", "reason": "your reason", "category": "out-of-context error"\}\},
\{\{"sentence": "third sentence", "reason": "your reason", "category": "entity error"\}\},
]
}

\vspace{2pt}
\noindent\textbf{Transcript:}

\noindent\texttt{\{transcript\}}

\vspace{2pt}
\noindent\textbf{Summary with \texttt{\{N\}} sentences:}

\noindent\texttt{\{summary\}}

\end{promptfigbox}

\caption{Prompt template for faithfulness evaluation.}
\label{fig:prompt_fact}
\end{figure*}
\begin{figure*}[t]
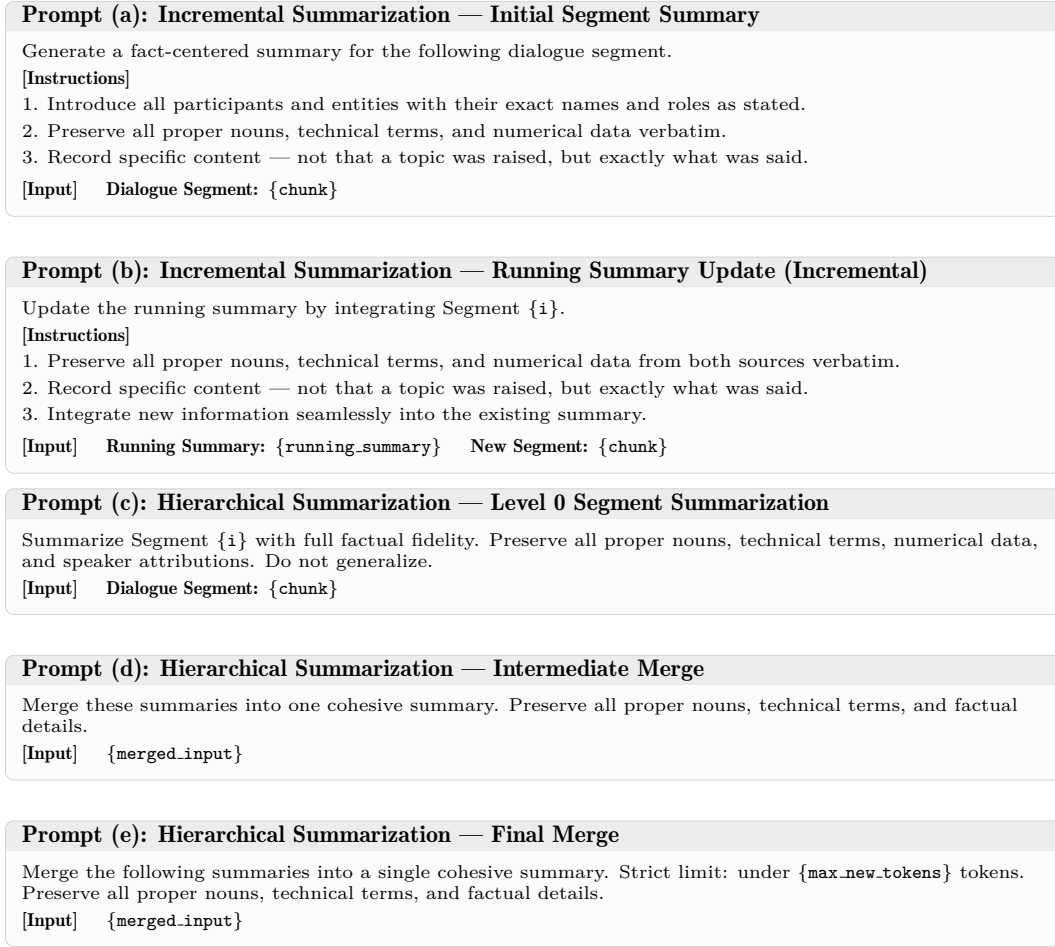

\centering

\begin{promptfigbox}{Prompt (a): Incremental Summarization --- Initial Segment Summary}

Generate a fact-centered summary for the following dialogue segment.

\vspace{2pt}
\noindent\textbf{[Instructions]}

\begin{promptenum}
\item Introduce all participants and entities with their exact
names and roles as stated.
\item Preserve all proper nouns, technical terms, and
numerical data verbatim.
\item Record specific content --- not that a topic was raised,
but exactly what was said.
\end{promptenum}

\vspace{2pt}
\noindent\textbf{[Input]} \quad \textbf{Dialogue Segment:} \texttt{\{chunk\}}

\end{promptfigbox}

\vspace{4pt}

\begin{promptfigbox}{Prompt (b): Incremental Summarization --- Running Summary Update (Incremental)}

Update the running summary by integrating Segment \texttt{\{i\}}.

\vspace{2pt}
\noindent\textbf{[Instructions]}

\begin{promptenum}
\item Preserve all proper nouns, technical terms, and
numerical data from both sources verbatim.
\item Record specific content --- not that a topic was raised,
but exactly what was said.
\item Integrate new information seamlessly into the existing
summary.
\end{promptenum}

\vspace{2pt}
\noindent\textbf{[Input]} \quad
\textbf{Running Summary:} \texttt{\{running\_summary\}} \quad
\textbf{New Segment:} \texttt{\{chunk\}}

\end{promptfigbox}

\begin{promptfigbox}{Prompt (c): Hierarchical Summarization --- Level 0 Segment Summarization}

Summarize Segment \texttt{\{i\}} with full factual fidelity.
Preserve all proper nouns, technical terms, numerical data, and speaker
attributions. Do not generalize.

\vspace{2pt}
\noindent\textbf{[Input]} \quad \textbf{Dialogue Segment:} \texttt{\{chunk\}}

\end{promptfigbox}

\vspace{4pt}

\begin{promptfigbox}{Prompt (d): Hierarchical Summarization --- Intermediate Merge}

Merge these summaries into one cohesive summary.
Preserve all proper nouns, technical terms, and factual details.

\vspace{2pt}
\noindent\textbf{[Input]} \quad \texttt{\{merged\_input\}}

\end{promptfigbox}

\vspace{4pt}

\begin{promptfigbox}{Prompt (e): Hierarchical Summarization --- Final Merge}

Merge the following summaries into a single cohesive summary.
Strict limit: under \texttt{\{max\_new\_tokens\}} tokens.
Preserve all proper nouns, technical terms, and factual details.

\vspace{2pt}
\noindent\textbf{[Input]} \quad \texttt{\{merged\_input\}}

\end{promptfigbox}

\caption{Prompt templates for summarization-based memory construction. \textbf{Incremental summarization}: (a) initialization generates the first summary from a new segment; (b) update integrates a new segment into an existing running summary. \textbf{Hierarchical summarization}: (c) level-0 summarizes each segment individually; (d) intermediate merge combines groups of segment summaries; (e) final merge produces the top-level summary under a strict token budget.}
\label{fig:prompt_summ_based}
\end{figure*}
\begin{figure*}[t]
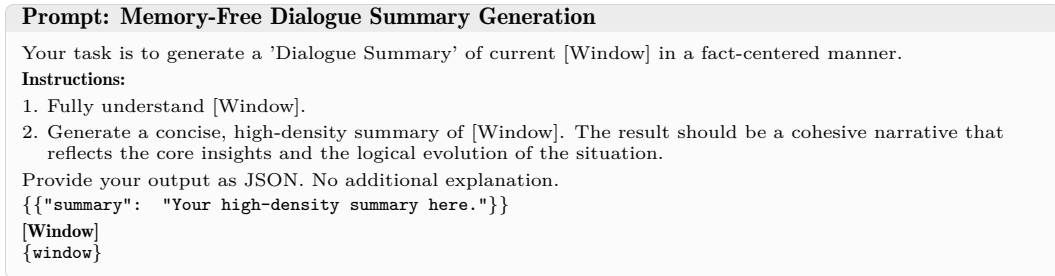

\centering
\begin{promptfigbox}{Prompt: Memory-Free Dialogue Summary Generation}

Your task is to generate a 'Dialogue Summary' of current [Window] in a fact-centered manner.

\vspace{2pt}
\noindent\textbf{Instructions:}

\begin{promptenum}
\item Fully understand [Window].
\item Generate a concise, high-density summary of [Window]. The result should be a cohesive narrative that reflects the core insights and the logical evolution of the situation.
\end{promptenum}

\noindent Provide your output as JSON. No additional explanation.

\vspace{1pt}
\noindent{\ttfamily\scriptsize
\{\{"summary": "Your high-density summary here."\}\}
}

\vspace{2pt}
\noindent\textbf{[Window]}

\noindent\texttt{\{window\}}

\end{promptfigbox}

\caption{Prompt template for dialogue summary generation.}
\label{fig:prompt_summary_generation}
\end{figure*}
\begin{figure*}[t]
\centering
\begin{promptfigbox}{Prompt: (i) Salient Utterance Identification and (ii) Contextual Dependence Judgment}

You are a Linguistic and Dialogue Context Analyst specializing in discourse coherence and pragmatics.
Your task is to analyze the provided dialogue excerpt to identify Salient Utterances and categorize their Contextual Gaps based ONLY on the provided text.

\vspace{2pt}
\noindent\textbf{\# Instructions}

\begin{promptenum}
\item Select sentences that contain a concrete decision, request, action item, or status update.
\begin{promptitems}
\item Merge Rule: If consecutive sentences form one coherent update/decision, merge them into a single entry.
\item Exclude: Simple acknowledgments ("Okay"), greetings, praise ("Great job"), or emotional reactions.
\end{promptitems}

\item For each salient utterance, judge whether it is contextually self-contained for summarization:
\begin{promptitems}
\item Identify pronouns, vague entities, underspecified concepts, missing attributes, and unstated reasoning that may prevent the utterance from being summarized independently.
\item Check whether the information needed to interpret the utterance in a stable and unambiguous way is explicitly available in the excerpt.
\item Assign a gap if the utterance is not sufficiently self-contained to support a contextually self-contained summary based on the current window alone.
\end{promptitems}

\item For each salient utterance, assign a gap type:
\begin{promptitems}
\item Coreference Resolution (C): a pronoun, shorthand reference, or vague noun phrase cannot be stably resolved from the excerpt.
\item Attribute/State Supply (A): a key concept, entity property, parameter, state, scope, or specification is missing, making the utterance insufficiently self-contained for summary writing.
\item Logic/Causal Connection (L): the utterance depends on prior reasoning, motivation, agreement, comparison, or constraint that is not given in the excerpt, so the conclusion or action is not fully supported within the window.
\item No Gap (N): the utterance is sufficiently self-contained to be summarized clearly and stably from the excerpt alone.
\end{promptitems}

\item Rules:
\begin{promptitems}
\item Use the standard of self-contained summarizability, not mere rough interpretability.
\item Mark N only when the utterance can be summarized from the current window without needing earlier dialogue to resolve its main meaning, reference, or rationale.
\item If nearby sentences fully resolve the missing information, do not mark a gap.
\item Do not mark a gap just because additional detail could make the utterance richer.
\item Commonly understood domain terms are not gaps by themselves, but project-specific concepts may be gaps if they remain underspecified for summary writing.
\item Clarification questions or examples can still have gaps if they rely on missing prior context.
\end{promptitems}
\end{promptenum}

\noindent Provide your answer in JSONL format. No additional explanation.

\vspace{1pt}
\noindent{\ttfamily\scriptsize
\{\{'sentence\_numbers': [number, number], 'sentences': ['text', 'text'], 'reason': 'concise gap analysis', 'gap\_type': 'N | C | A | L'\}\}
}

\vspace{1pt}
\noindent{\ttfamily\scriptsize
\{\{'sentence\_numbers': [number], 'sentences': ['text'], 'reason': 'concise gap analysis', 'gap\_type': 'N | C | A | L'\}\}
}

\vspace{2pt}
\noindent\texttt{---}

\vspace{1pt}
\noindent\textbf{\# [Input Dialogue]}

\noindent\texttt{\{window\}}

\end{promptfigbox}

\caption{Prompt template for salient utterance identification and contextual dependence judgment.}
\label{fig:prompt_salient}
\end{figure*}
\begin{figure*}[t]
\centering
\begin{promptfigbox}{Prompt: (iii) Candidate History Filtering}

You are a Context Retriever specializing in conversation threading.
Your task is to determine, for EACH chunk in [Dialogue Context], whether the chunk is relevant to the [Dialogue Window].
[Dialogue Context] contains multiple past-dialogue chunks.
You must evaluate each chunk independently and output one binary label per chunk.

\vspace{2pt}
\noindent A label should be:
\begin{promptitems}
\item 1: The chunk contains prior dialogue that is necessary or directly useful for understanding, continuing, or summarizing the current [Dialogue Window].
\item 0: The chunk is unrelated, only loosely topically similar, or not needed to maintain the conversational thread.
\end{promptitems}

\noindent A chunk is relevant if it contains at least one of the following:
\begin{promptenum}
\item A topic, entity, event, plan, task, or decision that is explicitly continued in the Dialogue Window.
\item Prior information needed to resolve references in the Dialogue Window, such as “this,” “that,” “it,” “the previous one,” “the method,” or similar expressions.
\item User preferences, constraints, assumptions, or prior decisions that directly affect the current window.
\item Earlier steps of the same ongoing task, project, or discussion.
\end{promptenum}

\noindent A chunk is NOT relevant if:
\begin{promptenum}
\item It discusses a different topic or a separate session.
\item It is only generally similar but not needed for the current window.
\item It contains background information that does not change the interpretation of the current window.
\item It requires guessing or inferring a connection that is not clearly supported.
\end{promptenum}

\noindent Important:
\begin{promptitems}
\item Evaluate each chunk separately.
\item Process chunks in 0-based index order, from the first chunk to the last chunk.
\item Do NOT skip any chunk, including the final chunk.
\item Before final output, verify that the list contains one label for every chunk index from 0 to N-1.
\item Do NOT assign one label to the entire Dialogue Context.
\item Do NOT rank the chunks.
\item Do NOT explain your reasoning.
\item Output only a Python-style list of 0s and 1s.
\item If there are N chunks in [Dialogue Context], your output must contain exactly N labels. Any output other than a single list of binary labels is invalid.
\end{promptitems}

\vspace{2pt}
\noindent\texttt{---}

\vspace{1pt}
\noindent\textbf{\# [Dialogue Window]}

\noindent\texttt{\{Dialogue\_Window\_Content\}}

\vspace{2pt}
\noindent\textbf{\# [Dialogue Context, 0-based index]}

\noindent\texttt{\{Dialogue\_Context\_Content\}}

\vspace{2pt}
\noindent\texttt{---}

\vspace{1pt}
\noindent\textbf{\# Output Format,}

\vspace{1pt}
\noindent{\ttfamily\scriptsize
[0,1,0,...]
}

\end{promptfigbox}

\caption{Prompt template for candidate history filtering.}
\label{fig:prompt_relevant_chunk_filtering}
\end{figure*}
\begin{figure*}[t]
\centering
\begin{promptfigbox}{Prompt: (iv) Evidence and Resolution Annotation, Part I}

You are a Contextual Gap Resolver. 
Your task is to audit EACH sentence in the [CURRENT SEARCH CHUNK] and decide whether the sentence provides explicit evidence that fills a concrete missing information slot in the [TARGET].

\vspace{2pt}
\noindent\textbf{Gap Type}

\begin{promptitems}
\item Coreference Resolution (C): a pronoun, shorthand reference, or vague noun phrase cannot be stably resolved from the excerpt.
\item Attribute/State Supply (A): a key concept, entity property, parameter, state, scope, or specification is missing, making the utterance insufficiently self-contained for summary writing.
\item Logic/Causal Connection (L): the utterance depends on prior reasoning, motivation, agreement, comparison, logic, cause, or constraint that is not given in the excerpt, so the conclusion or action is not fully supported within the window.
\end{promptitems}

\noindent You must act as a strict logic gate.

\begin{promptitems}
\item Use only explicit evidence in the [CURRENT SEARCH CHUNK].
\item Do not use outside knowledge.
\item Do not infer unstated links.
\item Do not mark a sentence as resolving the gap merely because it shares the same topic, entity, meeting, or general context.
\item A sentence resolves the gap \textbf{only if it explicitly provides information that the target depends on for independent understanding}, such as:
\end{promptitems}

\noindent A valid resolving sentence must explicitly provide at least one of the following:

\begin{promptitems}
\item the exact referent of a vague or ambiguous expression
\item a missing attribute, definition, state, scope, role, parameter, or specification
\item a necessary prior decision, rationale, agreement, comparison, condition, or event
\item an earlier action, request, or commitment that the target continues, refers back to, or presupposes
\end{promptitems}

\noindent Do not accept a sentence just because:

\begin{promptitems}
\item it is about the same topic
\item it mentions the same people, team, or project
\item it appears nearby in the discourse or event flow
\item it provides only broad background or loosely related context
\end{promptitems}

\end{promptfigbox}

\caption{Prompt template for evidence and resolution annotation, Part I.}
\label{fig:prompt_relevant_chunk-1}
\end{figure*}

\begin{figure*}[t]
\centering
\begin{promptfigbox}{Prompt: (iv) Evidence and Resolution Annotation, Part II}

\noindent\textbf{\# Strict Audit Protocol}

\begin{promptenum}
\item Read the [REFERENCE WINDOW], [TARGET], [Gap Type], and [Gap Reason], and determine the exact missing content needed to resolve the gap.
\begin{promptitems}
\item The missing content must be specific and concrete.
\item Do not define it as a broad topic, general context, or abstract category.
\end{promptitems}

\item Audit every sentence in the [CURRENT SEARCH CHUNK], from the first sentence.

\item For each sentence, think step-by-step:
\begin{promptitems}
\item Judge the sentence only with respect to the missing content implied by the [Gap Type] and [Gap Reason].
\item State whether the sentence provides explicit content that fills a concrete target gap.
\item Focus on what exact content the sentence provides or fails to provide, not on whether it is generally related.
\end{promptitems}

\item Write your judgment in \texttt{resolution\_reasoning}.
\begin{promptitems}
\item \texttt{resolution\_reasoning} must explain the sentence strictly in terms of its gap-resolution role.
\item If the sentence resolves the gap, briefly and concretely state what missing content it provides and how that helps make the target independently understandable.
\item If the sentence does not resolve the gap, state what the sentence does contain, then explain why that is still insufficient to resolve the gap.
\item Make clear whether the sentence is only topically related, background-only, or missing the actual dependency needed by the target.
\end{promptitems}

\item Only after writing \texttt{resolution\_reasoning}, decide the final status.
\begin{promptitems}
\item Mark the sentence as ``resolve'' only if it explicitly provides the missing content needed by the target.
\item Otherwise mark it as ``not\_resolve''.
\end{promptitems}
\end{promptenum}

\vspace{2pt}
\noindent\textbf{Additional constraints for \texttt{resolution\_reasoning}}

\begin{promptitems}
\item Do not say only that the sentence is “related,” “relevant,” “same context,” or “helpful.”
\item Do not justify using topic overlap, shared entities, or discourse proximity alone.
\item Do not restate the sentence without explaining its gap-resolution role.
\item Do not use vague phrases such as “this helps understand the context” unless you specify exactly what missing content it provides.
\item Keep it brief, but include the exact reason for resolve vs. not\_resolve.
\end{promptitems}

\noindent Provide your answer in JSONL format. No additional explanation.

\vspace{1pt}
\noindent{\ttfamily\scriptsize
\{\{"source\_id":"T0-S0","sentence":"...","reasoning":"...","status":"resolve"\}\}
}

\vspace{1pt}
\noindent{\ttfamily\scriptsize
\{\{"source\_id":"T0-S1","sentence":"...","reasoning":"...","status":"not\_resolve"\}\}
}

\vspace{2pt}
\noindent\textbf{[REFERENCE WINDOW] (Contextual Background)}

\noindent NOTE: This data is for contextual background only. Do NOT attempt to resolve the gap using this section, as it has been pre-verified that the required information is NOT present here.

\vspace{1pt}
\noindent\texttt{\{window\}}

\vspace{2pt}
\noindent\textbf{[TARGET]}

\noindent\textbf{[Sentence]:} \texttt{\{target\_sentence\}}

\noindent\textbf{[Gap Type]:} \texttt{\{gap\_type\}}

\noindent\textbf{[Gap Reason]:} \texttt{\{gap\_reason\}}

\vspace{2pt}
\noindent\textbf{[CURRENT SEARCH CHUNK]}

\noindent\texttt{\{search\_chunk\}}

\end{promptfigbox}

\caption{Prompt template for evidence and resolution annotation, Part II.}
\label{fig:prompt_relevant_chunk-2}
\end{figure*}
\begin{figure*}[t]
\centering
\begin{promptfigbox}{Prompt: Critical Logic Audit for Contextual Gap Resolution}

You are a Critical Logic Auditor. Your task is to perform a final validation of a "Contextual Gap Resolution" result. You must determine if the provided [RESOLVING SENTENCE] truly and sufficiently fills the [GAP] in the [TARGET SENTENCE] as claimed in the [REASONING].

\vspace{2pt}
\noindent\textbf{\# Validation Criteria:}

\begin{promptenum}
\item Sufficiency: Does the resolving sentence provide the \emph{exact} missing information (referent, attribute, or logic) defined in the Gap Reason?
\item Non-Inference: Is the resolution explicit? If you still need to "guess" or "assume" anything to make the target clear, the resolution is INVALID.
\item Independence: After incorporating the information from the resolving sentence, can the target sentence now be summarized as a standalone, unambiguous fact?
\item Redundancy: Is the information in the resolving sentence meaningfully distinct from what is already present in the [REFERENCE WINDOW]? If the resolving sentence merely restates or duplicates content already available in the window, the resolution is INVALID --- it provides no new contextual value beyond what was already accessible.
\end{promptenum}

\vspace{2pt}
\noindent\textbf{\# Evaluation Rules:}

\begin{promptitems}
\item 1: The evidence is explicit, directly addresses the gap, and makes the target self-contained.
\item 0: The reasoning is based on topical overlap, the evidence is too vague, or the resolution requires additional unstated context.
\end{promptitems}

\vspace{2pt}
\noindent\textbf{\# Instructions:}

\begin{promptitems}
\item Audit each entry in the [CANDIDATE LIST] independently.
\item Identify the "source\_id" for every candidate that receives a score of \textbf{1}.
\item Return the results \textbf{strictly} as a JSON list of strings containing only the successful "source\_id"s. 
\item No preamble, no explanation, no additional text.
\end{promptitems}

\vspace{2pt}
\noindent\textbf{\# Output format:}

\vspace{1pt}
\noindent{\ttfamily\scriptsize
["Tn-Sm", "Tn-Sm", ...]
}

\vspace{2pt}
\noindent\textbf{[REFERENCE WINDOW] (Contextual Background)}

\noindent NOTE: This data is for contextual background only. Do NOT attempt to resolve the gap using this section, as it has been pre-verified that the required information is NOT present here.

\vspace{1pt}
\noindent\texttt{\{window\}}

\vspace{2pt}
\noindent\textbf{[TARGET]}

\noindent\textbf{[SENTENCE]} \texttt{\{target\_sentence\}}

\noindent\textbf{[GAP TYPE]} \texttt{\{gap\_type\}}

\noindent\textbf{[GAP REASON]} \texttt{\{gap\_reason\}}

\vspace{2pt}
\noindent\textbf{[CANDIDATE LIST]}

\noindent\texttt{\{candidate\}}

\end{promptfigbox}

\caption{Prompt template for final validation of contextual gap resolution annotations.}
\label{fig:prompt_relevant_chunk_validation}
\end{figure*}
\end{document}